\pdfoutput=1
\documentclass[11pt]{article}

\usepackage[preprint]{acl}

\usepackage{times}
\usepackage{siunitx} 
\usepackage{latexsym}
\usepackage{tablefootnote}
\usepackage{threeparttable}
\usepackage{pifont} 
\usepackage{verbatim}
\usepackage{tcolorbox}
\usepackage[T1]{fontenc}

\usepackage[utf8]{inputenc}

\usepackage[verbose]{microtype}

\usepackage{inconsolata}
\usepackage{amsmath}
\usepackage{array}
\usepackage{multirow}
\usepackage{kotex}
\usepackage{adjustbox}
\usepackage{url}
\usepackage{xcolor}
\usepackage{multicol}
\usepackage{ulem} 
\newcommand\highlight[1][yellow]{%
  \bgroup 
  \markoverwith{\textcolor{#1}{\vrule width.1em height.8em depth.2em}}%
  \ULon 
}
\usepackage{booktabs}
\usepackage{makecell}
\usepackage{tabularx}
\usepackage{graphicx}
\usepackage{color,soul}
\usepackage{lscape}

\title{DSLR: Document Refinement with Sentence-Level Re-ranking and Reconstruction to Enhance Retrieval-Augmented Generation}

\author{Taeho Hwang
        \quad Soyeong Jeong
        \quad Sukmin Cho
        \quad SeungYoon Han
        \quad Jong C. Park\thanks{\hspace{0.2cm}Corresponding author} \\
        School of Computing \\
        Korea Advanced Institute of Science and Technology\\ 
       \texttt{\{doubleyyh, starsuzi, nelllpic, seungyoonee, jongpark\}@kaist.ac.kr}}

\begin{document}
\maketitle
\begin{abstract}
Recent advancements in Large Language Models (LLMs) have significantly improved their performance across various Natural Language Processing (NLP) tasks.
However, LLMs still struggle with generating non-factual responses due to limitations in their parametric memory.
Retrieval-Augmented Generation (RAG) systems address this issue by incorporating external knowledge with a retrieval module.
Despite their successes, however, current RAG systems face challenges with retrieval failures and the limited ability of LLMs to filter out irrelevant information.
Therefore, in this work, we propose \textit{\textbf{DSLR}} (\textbf{D}ocument Refinement with \textbf{S}entence-\textbf{L}evel \textbf{R}e-ranking and Reconstruction), an unsupervised framework that decomposes retrieved documents into sentences, filters out irrelevant sentences, and reconstructs them again into coherent passages.
We experimentally validate \textit{DSLR} on multiple open-domain QA datasets and the results demonstrate that \textit{DSLR} significantly enhances the RAG performance over conventional fixed-size passage.
Furthermore, our \textit{DSLR} enhances performance in specific, yet realistic scenarios without the need for additional training, providing an effective and efficient solution for refining retrieved documents in RAG systems.
\end{abstract}

\section{Introduction}

Recent advancements in Large Language Models (LLMs) \cite{fewshotlearner, GPT-4_technical_report, Llama2} have significantly expanded their capabilities across diverse knowledge-intensive tasks in Natural Language Processing (NLP), such as Question Answering (QA) \cite{NQ, TQA, SQD}. 
However, despite these capabilities, LLMs still face challenges such as generating plausible yet non-factual responses, known as hallucination, due to their reliance on limited parametric memory \cite{DBLP:conf/acl/MallenAZDKH23}. 
Also, it is noted that this parametric memory is static, as LLMs can learn knowledge only up to the specific date on which the training was completed.
Therefore, these limitations restrict their adaptability to long-tailed or ever-evolving domains \cite{Realtime} and to unseen knowledge outside their training data \cite{kalmv}.

\begin{figure*}
    \centering
    \includegraphics[width=\textwidth]{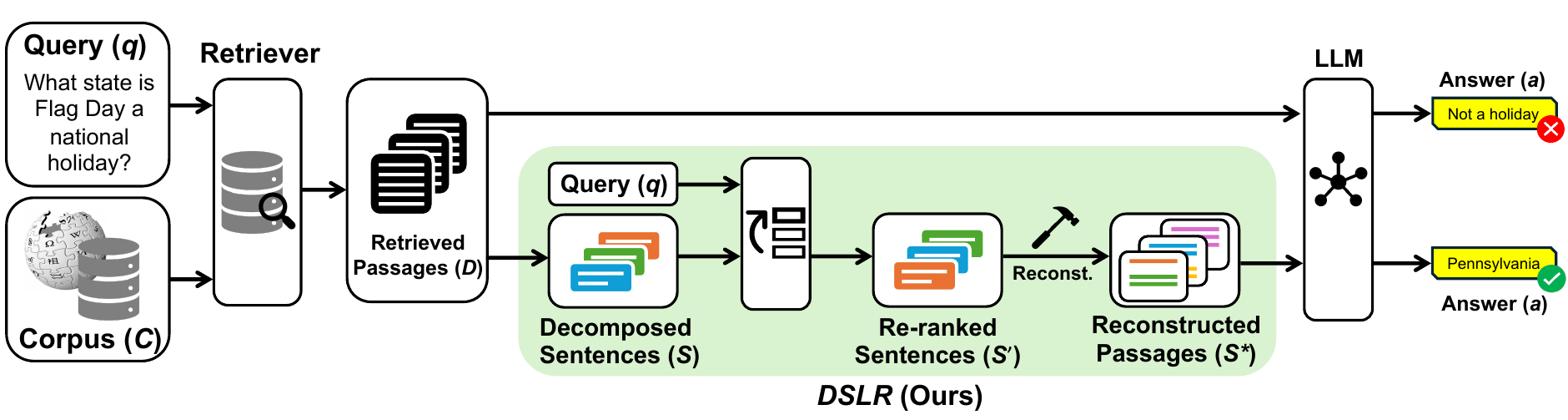}
    \caption{
\small Comparison of the conventional RAG pipeline (top) and our sentence-level re-ranking and reconstruction framework (bottom) in an RAG system. Initially, both methods retrieve query-relevant documents at the passage level. The conventional approach directly utilizes these passages, which may contain redundant information leading to QA inaccuracies. By contrast, our method decomposes passages into sentences, re-ranks them based on relevance, and reconstructs them into coherent passages for more accurate LLM responses.}
    \label{fig:dslr}
    \vspace{-0.125in}
\end{figure*}

Retrieval-Augmented Generation (RAG) \cite{DBLP:conf/iclr/KhandelwalLJZL20, RAG, DBLP:conf/icml/BorgeaudMHCRM0L22, replug} has been introduced as an effective solution to address such problems.
Specifically, RAG enhances LLMs by integrating non-parametric memories fetched from external knowledge bases using a retrieval module, which helps LLMs' responses grounded on factual evidence and makes them more up-to-date. 

While the efficacy of RAG depends on the performance of the retrieval module, the instability of LLMs in incorporating the retrieved knowledge is also a critical challenge to RAG.
To be specific, retrieved documents sometimes contain irrelevant information~\cite{das}, and LLMs often struggle to effectively filter out such redundant details and focus on the most query-relevant knowledge~\cite{distractor, distractor2, lostinthemiddle, distractor3}, which leads to the failure of the overall RAG systems.
Therefore, it is crucial to investigate how to effectively refine retrieved documents before augmenting them with LLMs, ensuring that the LLMs are not distracted by irrelevant information within retrieved documents.

Re-ranking the order of the retrieved document set~\cite{monot5, llm-rerank} or refining them into new documents~\cite{filco,recomp} can be considered as solutions. 
However, they generally require high computational costs for training additional re-ranking or refining models.
Another proposed solution is to reduce the retrieval granularity from passage-level to sentence-level which can help eliminate redundant information within passages \cite{phrase_retriever, denseXretrieval}.
However, this might also inadvertently remove important contextual information, which is crucial for accurately answering the given queries \cite{decontextualization}.
Therefore, we should explore a novel method that can effectively and efficiently filter out irrelevant information while maintaining the necessary contextual details.

In this work, we introduce an unsupervised \textbf{\textit{DSLR}} (\textbf{D}ocument Refinement with \textbf{S}entence-\textbf{L}evel \textbf{R}e-ranking and Reconstruction) framework that consists of three steps: 1) decomposition, 2) re-ranking, and 3) reconstruction.
Specifically, after retrieving the passage-level document, the \textit{DSLR} framework operates by first decomposing the retrieved document into sentences for finer granularity and then filtering out the irrelevant sentences based on their re-ranking scores from the ranking models, including off-the-shelf retrievers and re-rankers. 
Finally, the remaining sentences are reconstructed into a single document to preserve the original contextual information. 
Note that \textit{DSLR} is an unsupervised refinement framework, which does not require any additional training for re-ranking or reconstruction steps.
The overall \textit{DSLR} framework is illustrated in Figure~\ref{fig:dslr}.

We validate our framework across a diverse range of open-domain QA benchmarks, which include three general QA datasets and three specific QA datasets that require domain-specific or ever-evolving knowledge.
Our experimental results show that \textit{DSLR} significantly enhances the overall RAG performance and is comparable to, or even outperforms, the supervised baseline approaches.
Specifically, when evaluated with specific QA datasets, \textit{DSLR} shows high robustness in realistic settings.
Furthermore, a detailed analysis demonstrates the effectiveness of each proposed step and how it contributes to the overall performance.

Our contributions in this work are threefold:
\vspace{-0.075in}
\begin{itemize}
  \item We point out that recent RAG systems are largely vulnerable to redundant knowledge within fixed-size passage-level retrieved documents and that the existing refining strategies generally require additional training steps.
  \item We propose a \textit{DSLR} framework that incorporates sentence-level re-ranking and reconstruction to effectively remove redundant knowledge that negatively affects the RAG system.
  \item We show that \textit{DSLR} is highly effective and efficient even without additional training steps in both general and specific scenarios.
\end{itemize}

\section{Related Work}





\paragraph{Information Retrieval.}
Information Retrieval (IR) is the task of searching for query-relevant documents from a large corpus \cite{IR}, which has been widely applied for both search systems and various NLP tasks such as open-domain QA ~\cite{KILT}.
IR models can be categorized into sparse retrievers \cite{tfidf, bm25}, which use lexical metrics to calculate relevance scores between queries and documents, and dense retrievers \cite{DPR-karpukhin, contriever}, which embed queries and documents into a dense space that captures semantic relationships but requires significant computational resources \cite{dar}.
 

In order to further enhance retrieval performance, additional strategies have been proposed.
Specifically, the re-ranking strategy improves retrieval performance by recalculating relevance scores using an additional re-ranking model \cite{bert-rerank, monot5, rankt5}, and then reordering the documents based on these scores.
Recently, LLMs have shown remarkable re-ranking performance by generating relevance labels without requiring further fine-tuning \cite{holistic, prp}.

While the aforementioned work on IR \cite{100words, DPR-karpukhin} generally assumes fixed-size, 100-word passages as the document length, some work has explored an optimal level of retrieval granularity~\cite{phrase, phrase_retriever,proconvqa,denseXretrieval}.
These approaches validate that a fine-grained level of granularity, containing only the knowledge needed to answer the query, can enhance the overall performance by excluding redundant details in the lengthy retrieved documents. 
However, reducing retrieval granularity to the sentence level can disrupt the original context and result in a loss of the document’s coherence \cite{decontextualization}.
In addition, sentence-level retrieval generally requires a much larger index size compared to passage-level retrieval \cite{DBLP:conf/emnlp/LeeWC21}.
By contrast, we investigate a novel framework for effectively re-ranking sentences within retrieved passage-level documents and then reconstructing the re-ranked sentences to preserve contextual integrity.

\paragraph{Retrieval-Augmented Generation.} RAG has emerged as a promising solution for addressing LLMs' hallucination issues by leveraging external knowledge fetched by the retrieval module. 
Specifically, RAG incorporates retrieval modules that reduce the need to update the parameters of LLMs and help them generate accurate and reliable responses~\cite{DBLP:conf/iclr/KhandelwalLJZL20, RAG, DBLP:conf/icml/BorgeaudMHCRM0L22, replug}. 
Additionally, various real-world applications integrate RAG as a core component when deploying LLM-based services~\cite{plugin, langchain, Toolllm}. 
However, they still have limitations due to the imperfections of the retrieval module within RAG, where the retrieved documents containing query-irrelevant information can negatively lead the LLMs to generate inaccurate answers.


To address them, several studies have attempted to leverage the capabilities of LLMs to enhance their resilience against irrelevant knowledge. 
These approaches include crafting specialized prompts~\cite{self-ask, das}, training plug-in knowledge verification models~\cite{kalmv}, adaptively retrieving the required knowledge~\cite{adaptive-rag, selfrag, ReFeed}, and augmenting knowledge using the capabilities of the LLM itself \cite{gen_read}.
Among the promising solutions, recent studies show that further refining the retrieved documents into fine-grained knowledge can improve the RAG performance~\cite{recomp, REAR, filco, bider}.
However, such refinement strategies generally require additional fine-tuning on a specific dataset, which might result in limited generalizability and high computational cost. 
By contrast, our proposed refinement framework removes irrelevant information with unsupervised sentence-level re-ranking and reconstruction steps by using off-the-shelf ranking models without requiring additional training costs.

\section{Method}

In this section, we describe a novel framework \textit{DSLR} for enhancing the precision of retrieval results through sentence-level ranking and reconstruction, integrated into the RAG system. Note that \textit{DSLR} does not require additional training.


\subsection{Preliminaries}

\begin{figure}
    \centering
    \includegraphics[width=\columnwidth]{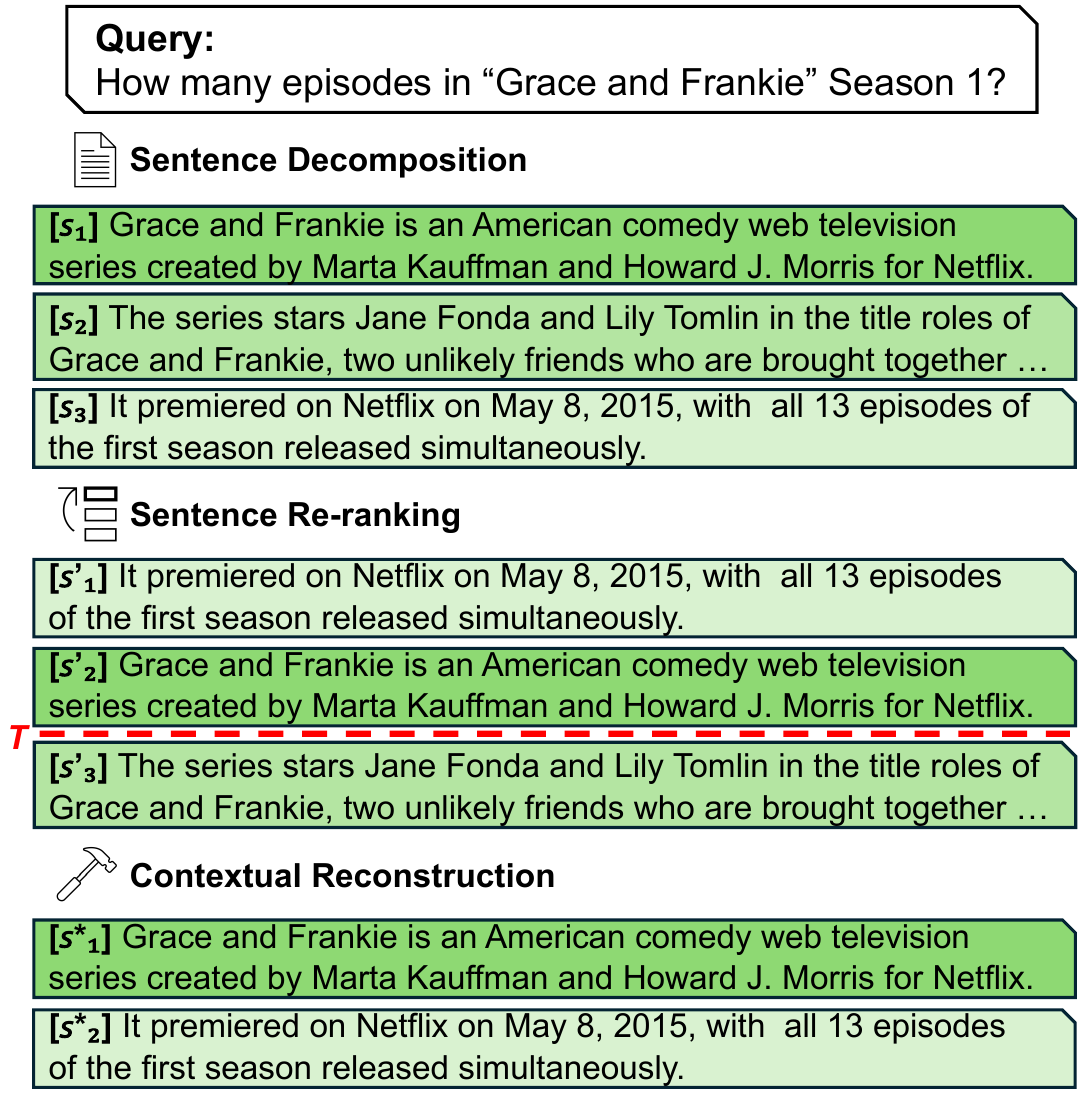}
    \caption{
\small Examples of each step in the \textit{DSLR} framework, which consists of three steps: 1) Sentence Decomposition 2) Sentence Re-ranking, and 3) Contextual Reconstruction.}
    \label{fig:method}
    \vspace{-0.125in}
\end{figure}
We first introduce the general RAG system, which consists of three steps: the retrieval step, the re-ranking step, and the generation step.
Note that all steps focus on passage-level documents.


\subsubsection{Retrieval Step}
\label{Retrieval Step}
The retrieval step searches for a potentially relevant document set $\mathcal{D}$ to the given query $q$ from a retrieval corpus $\mathcal{C}$ consisting of millions of documents. 
This retrieval step is conventionally performed using a sparse retriever $S$, such as BM25, which is widely used for processing large corpora due to its low latency. 
The sparse retriever $S$ fetches the relevant documents having high relevant scores based on lexical values such as document length or unique word count.
Formally, we define the retrieval step as:
\[
\mathcal{D} = \text{Retrieve}(q, \mathcal{C}; S) = \{d_1, d_2, ..., d_n\}
\]
where $d_k$ represents a document having the top-$k$ score among the retrieval corpus $\mathcal{C}$ for a given query $q$, and $n$ denotes the size of $\mathcal{D}$, generally ranging from tens to hundreds.

\subsubsection{Re-ranking Step}\label{Re-ranking step}

While the sparse retriever $S$ can efficiently handle a large corpus, it cannot consider semantic similarities, thereby limiting its retrieval performance for lexically different but semantically relevant pairs. 
To address this, the re-ranking step aims for more precise retrieval results by reordering the retrieved document set $\mathcal{D}$ using the ranking model $R$. 
This model transforms $\mathcal{D}$ into a newly ordered document set $\mathcal{D'}$ based on relevance scores with a query $q$, capturing semantic meanings that could not be addressed in the retrieval step with $S$.
Formally, we define the re-ranking step as:
\[
\mathcal{D'}=\text{Re-rank}(q,\mathcal{D};R)=\{d'_1,\ldots,d'_m\}
\]
where $d'_k$ represents the document that has top-$k$ relevance score among $\mathcal{D}$ and \(m \ll n\), indicating that the subset $\mathcal{D'}$ contains significantly fewer documents than the original set $\mathcal{D}$.







\subsubsection{Generation Step}
\label{Generation Step}
After the re-ranking step, the document set $\mathcal{D'}$ is augmented to the LLM $M$ with the supporting documents to generate the correct answer $a$ for the given query $q$. The generation step can be formalized as:
\[
 a=\text{Generate}(q, \mathcal{D'}; M)
\]

In RAG systems, the three key steps are designed to retrieve the most query-relevant knowledge for LLMs, typically at the passage level. However, this fixed granularity can overlook finer relevance between queries and individual sentences.
Therefore, in this work, we introduce a fine-grained, sentence-level ranking strategy in the re-ranking step, aiming to reduce distractions from irrelevant information and enhance answer accuracy.



\subsection{Document Refinement with Sentence-Level Re-ranking and Reconstruction (DSLR)}

We propose a novel unsupervised refinement framework, \textit{D}ocument Refinement with \textit{S}entence-\textit{L}evel \textit{R}e-ranking and Reconstruction (\textit{DSLR}), designed to assess the fine-grained relevance of individual sentences within a passage and reconstruct to preserve the original contextual coherence. 
Figure \ref{fig:method} illustrates examples generated by each step in our \textit{DSLR} framework.


\subsubsection{Sentence Decomposition and Re-ranking}
After the retrieval step (§\ref{Retrieval Step}), we conduct sentence-level re-ranking for the documents within the retrieved set $\mathcal{D}$. First, each document $d_i \in \mathcal{D}$ is decomposed into a sentence set $\mathcal{S}_i = \{s_j\}_{j=1}^l$, where $s_j$ represents the $j$-th sentence in document $d_i$ and $l$ is the number of sentences in $d_i$. Then, the passage-level retrieved set $\mathcal{D}$ is redefined to the sentence-level retrieved set $\mathcal{S} = \cup_{i=1}^n \mathcal{S}_i$. For instance, as illustrated in Figure \ref{fig:method}, a passage retrieved for a query ``How many episodes in "Grace and Frankie" Season 1?" is decomposed into three sentences \( s_1 \), \( s_2 \), and \( s_3 \) during the sentence decomposition step.

To extract sentences containing relevant information for a query \( q \), we initially perform re-ranking to assess relevance scores at the sentence level. Sentences in \( \mathcal{S} \) with scores below a predefined threshold \( T \) are deemed irrelevant and removed, resulting in a refined set \( \mathcal{S'} \). The sentence-level re-ranking is formally defined as follows:
\[
\mathcal{S'}=\text{Re-rank}(q,\mathcal{S};R)=\{s'_1,\ldots,s'_m\}
\]
where each $s'_k$ is a sentence from $\mathcal{S}$ whose relevance score exceeds $T$. Figure~\ref{fig:method} demonstrates the reordering of sentences, highlighting the exclusion of $s'_3$ due to its insufficient relevance score.
Note that this step of the \textit{DSLR} framework utilizes off-the-shelf ranking models, which are identical to those used in passage-level re-ranking.



\subsubsection{Contextual Reconstruction}

While the sentence decomposition and re-ranking steps select the top-$m$ relevant sentences for the query $q$, these sentences may lack contextual relationships to one another, as these steps can disrupt the original contextual flow of the passage by discarding some sentences.
Instead of following a widely used approach of simply concatenating these sentences in descending order of their relevance scores, we propose to reconstruct them into the contextually organized set, $\mathcal{S}^*$, to reflect the order in which they were originally positioned before being decomposed from passages, ensuring the original coherence and logical flow:
\[
\mathcal{S}^* = \text{Reconstruction}(\mathcal{S'},\mathcal{S})=\{s^*_1,\ldots,s^*_m\}
\]
where $s^*_i$ is the sentence included in $S'$ and $i$ denotes the relative position of $s^*_i$ within $\mathcal{S}$. 
As shown in Figure~\ref{fig:method}, the remaining two sentences are reconstructed in their original order by switching their positions to preserve the context before the sentence re-ranking step.
Then, LLM $M$ generates the answer $a$ for a given query $q$ with $\mathcal{S}^*$ formalized as: $a=\text{Generate}(q,\mathcal{S}^*;M)$.

\begin{table*}[t!]
\vspace{-0.1in}
\small
\centering

\begin{threeparttable}
\renewcommand{\thefootnote}{\fnsymbol{footnote}} 
\resizebox{\textwidth}{!}{%
\begin{tabular}{ll | cc | cc | cc | cc | cc | cc | cc}
\toprule 
\multirow{2}{*}{\textbf{Type}}    & \multirow{2}{*}{\textbf{Re-ranker}}            & \multicolumn{2}{c|}{\textbf{NQ}}            & \multicolumn{2}{c|}{\textbf{TQA}}           & \multicolumn{2}{c|}{\textbf{SQD}}           & \multicolumn{2}{c|}{\textbf{RQA}}           & \multicolumn{2}{c|}{\textbf{SQ}} & \multicolumn{2}{c|}{\textbf{BASQ}} & \multicolumn{2}{c}{\textbf{AVG.}}            \\ 

                              &                    & \# tok& Acc & \# tok& Acc & \# tok& Acc & \# tok& Acc & \# tok& Acc & \# tok& Acc & \# tok& Acc \\ \midrule 
\multicolumn{16}{c}{\textit{Baseline}}   \\  \midrule  \midrule
 - & -   & 167& 25.6& 170& 58.0& 166& 28.5& 1277& 41.1& 162& 33.9& 444& 56.7& 398& 40.6\\  \midrule
 \multicolumn{16}{c}{\textit{Ours}}   \\  \midrule \midrule
             Sparse Ret.                &                   BM25                &    48& 28.7& 81& 60.8& 41& 28.0& 689& 40.4& 52& 40.7& 202& 52.6& 186& 41.9\\ \midrule
\multirow{2}{*}{Dense Ret.}       & {Contriever} 
                                                 & 68& 29.2& 60& 62.0& 61& 29.1& 418& 41.2& 69& 40.8& 308& 57.2& 164& 43.2\\ 
                              &  DPR          & 61& 33.6& 74& 62.9& 56& 27.3& 517& 40.1& 75& 40.9& 309& 55.9& 182& 43.4\\ \midrule
\multirow{2}{*}{Supervised Re-r.}   & MonoT5      & 74& 31.1& 84& 62.3& 67& \textbf{30.4}& 625& 42.1& 50& \textbf{41.1}& 363& 57.2& 211& 44.0\\
                              & RankT5           & 83& 29.4& 69& 61.7& 60& \textbf{30.4}& 475& 41.6& 49& 40.6& 337& 57.2& 179& 43.5\\ \midrule
Unsupervised Re-r.  & RG  & 46& \textbf{33.7} & 76& \textbf{64.1} & 51& 29.5& 534& \textbf{42.5}& 97& 38.9& 291& \textbf{59.5}& 183& \textbf{44.7}\\ \bottomrule
\end{tabular}%
}


\end{threeparttable}

\caption{\small Performance comparison between the \textit{Baseline} (original top-1 document) and \textit{Ours} (\textit{DSLR}-refined top-1 document) on various open-domain QA datasets. The table shows the average token count (\# tok) and accuracy (Acc) for both sparse and dense retrievers, as well as for supervised and unsupervised re-rankers. Best results are in \textbf{bold}.}
\label{tab:main1}
\end{table*}

\section{Experiment Setups}

In this section, we describe the experimental setup for evaluating \textit{DSLR} across various scenarios. We provide additional details in Appendix \ref{sec:Experimental_Setups}.


\subsection{Models}

\noindent\textbf{Retriever.} We use BM25~\cite{bm25} as a passage-level retriever, which is a widely used sparse retriever due to its notable performance with high efficiency. The retriever fetches the \textbf{top-1} passage-level query-relevant document from an external corpus, which serves as the baseline document.

\noindent\textbf{Re-ranker.} 
We operationalize a variety of ranking models as re-rankers, including off-the-shelf retrievers, fine-tuned re-rankers, and LLMs.
\textbf{1) Sparse Retriever:} We use \textbf{BM25}~\cite{bm25} as a sentence-level re-ranker. Note that BM25 is only applied at the sentence level, as it is primarily utilized in the retrieval step.
\textbf{2) Dense Retriever:} We utilize two representative dense retrievers, \textbf{Contriever}~\cite{contriever} and \textbf{DPR}~\cite{DPR-karpukhin}, which are better at capturing the semantic similarity between documents and queries than sparse retrievers.
\textbf{3) Supervised Re-ranker\footnote{It is important to note that the terms `supervised' and `unsupervised' in this context refer to the models being trained on document ranking tasks, and not on document refinement tasks.\label{footnote}}:} We employ two supervised re-ranking models based on T5~\cite{T5}, \textbf{MonoT5}~\cite{monot5} and \textbf{RankT5}~\cite{rankt5}. These models are specifically trained for pointwise document ranking tasks.
\textbf{4) Unsupervised Re-ranker\footref{footnote}:}
We explore \textbf{Relevance Generation (RG)}~\cite{holistic}, a pointwise ranking method using the inherent ranking ability of LLMs, validating its effectiveness in scenarios lacking extensive labeled data.
We use LLama2-13b-chat~\cite{Llama2} as a ranking model for \textbf{RG}.

\noindent\textbf{Reader.}
We use the instruction-tuned, open-source LLM \textbf{LLama2-13b-chat} as our reader. To generate the final answer, the document is prepended to the system prompt.

\subsection{Datasets}

We evaluate our \textit{DSLR} across 6 open-domain QA datasets, including both general and specific domains.
First, we conduct our experiment using the development set of \textbf{Natural Questions (NQ)}~\cite{NQ}, \textbf{TriviaQA (TQA)}~\cite{TQA}, and \textbf{SQuAD (SQD)}~\cite{SQD}, consisting of queries with general topics.
Additionally, we incorporate specialized datasets such as \textbf{RealtimeQA (RQA)}~\cite{Realtime}, \textbf{SciQ (SQ)}~\cite{sciq}, and \textbf{BioASQ (BASQ)}~\cite{bioasq, bioasq2} for evaluating the generalizability of our proposed method.
In detail, RQA includes questions that are updated periodically to test our system's ability to handle ever-evolving knowledge. 
In addition, SQ and BASQ are domain-specific datasets in science and biology, respectively.
Specifically, for BASQ, we selectively use the questions from the BioASQ6 challenge (task b) that are suitable for yes/no and factoid responses.
We report the effectiveness of our framework with \textbf{Accuracy (Acc)}, which determines whether the prediction contains golden answers, following~\citet{selfrag}. 

\begin{figure*}[t!]
    \centering
    \includegraphics[width=1\linewidth]{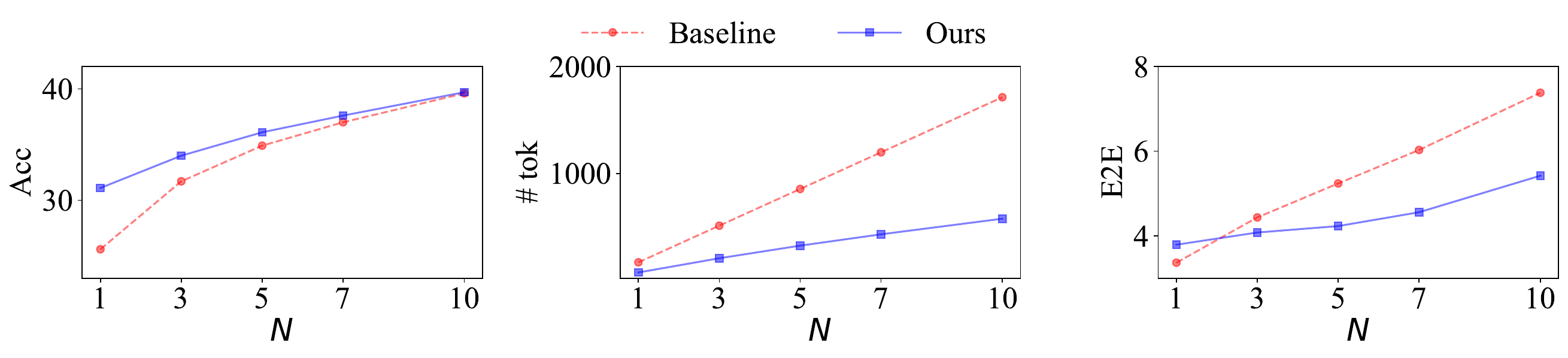}
    \caption{\small Comparison between the Baseline (original  documents) and Ours (\textit{DSLR}-refined documents using MonoT5) in the top-\(N\) multiple passages scenario on the NQ dataset. (Left) Accuracy (Acc) as top-\(N\) increases. (Center) Average token count (\# tok) as top-\(N\) increases. (Right) Average end-to-end latency (E2E) as top-\(N\) increases, measured in seconds.}
    \label{fig:various}
    \vspace{-0.15in}
\end{figure*}

\subsection{Implementation Details}


The threshold \( T \), used to remove irrelevant content, was determined empirically by sampling 1,000 random entries from each of the NQ, TQA, and SQD training sets and setting \( T \) to the relevance score at the 90th percentile. Detailed values of \( T \) for various models are provided in Table \ref{tab:public_threshold}.
The retrieval corpus for NQ, TQA, and SQD is a pre-processed Wikipedia dump from Dec. 20, 2018 following~\citet{DPR-karpukhin}, and for BASQ and RQA, we use their own retrieval corpora.
To be specific, BASQ used the BEIR (v1.0.0)~\footnote{\url{https://github.com/beir-cellar/beir}} BioASQ corpus, specializing in biomedical information retrieval.
For the RQA dataset, spanning from 2022 to 2023, we use the search documents provided at the time of dataset creation through the Google Cloud Search (GCS) API to align the periods of the queries and answers. 
When implementing each component in \textit{DSLR}, we decompose passage-level documents into sentences using the Sentencizer from Spacy\footnote{\url{https://spacy.io/}}. All predictions in our experiments are generated via greedy decoding. 



\section{Experimental Results and Analyses}
In this section, we show the overall experimental results with in-depth analyses of our framework. 

\paragraph{Main Results.} 

First of all, Table \ref{tab:main1} shows that our \textit{DSLR}-refined top-1 document consistently outperforms the original top-1 document across all datasets and scenarios, despite reduced token counts. This confirms our hypothesis that the redundant information within the fix-sized passages adversely affects the RAG performance and highlights the importance of providing only query-relevant information in RAG with finer-grained sentences.

Furthermore, \textit{DSLR} also shows performance enhancement over specialized datasets, such as ever-evolving RQA and domain-specific SQ and BASQ datasets.
Specifically, the re-rankers based on pre-trained models such as T5 and the LLM demonstrate remarkable performance improvement.
Given that \textit{DSLR} requires no additional training, the robust and effective performance suggests its applicability to diverse real-world scenarios, particularly where queries frequently change across different timelines and domains.

\paragraph{\textit{DSLR} in Multiple Passages.}
To assess the effectiveness and efficiency of \textit{DSLR} in multiple passages, we gradually increased the number of documents \(N\) and compared the performance, token count, and end-to-end (E2E) latency\footnote{These experiments were conducted using four V100 GPUs.} of the original top-\(N\) documents with those refined by \textit{DSLR}. 

As shown in the left panel of Figure \ref{fig:various}, both sets of documents show consistent performance improvements as \(N\) increases. However, \textit{DSLR} consistently outperforms the original documents across all \(N\) levels, with more notable differences at lower \(N\) values. This suggests that \textit{DSLR} can significantly enhance performance in RAG, even as the number of documents increases.

Due to the quadratic increase in memory and time requirements with the number of tokens in transformer-based LLMs, reducing the token count is crucial for improving efficiency \cite{attention}. As depicted in the center and right panels of Figure \ref{fig:various}, \textit{DSLR} substantially reduces the token count compared to the original documents, with the difference becoming more significant as \(N\) increases. This reduction in tokens also decreases E2E latency in all scenarios except top-1. Notably, at top-10, while the performance difference is minimal (39.6 vs. 39.7), the token count reduction from 1,713 to 577 (nearly 2.97 times) and the corresponding E2E latency reduction from 7.382 seconds to 5.422 seconds (nearly 2 seconds) demonstrate that \textit{DSLR} can enhance both performance and efficiency in RAG. Detailed results are available in Table \ref{tab:detail_N}.

\begin{figure}[t!]
    \centering
    \includegraphics[width=1\linewidth]{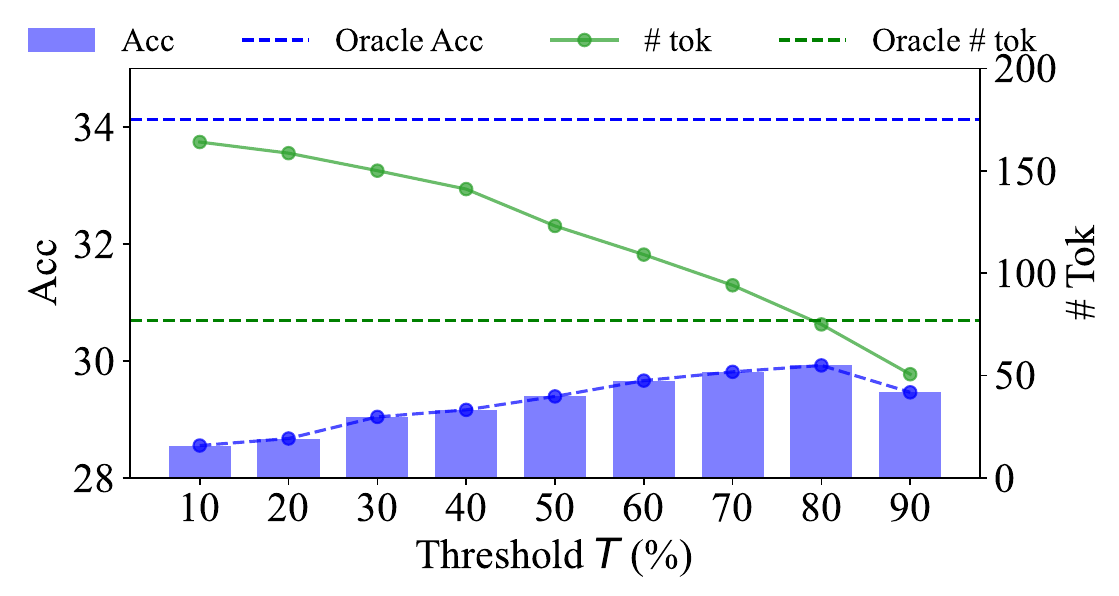}
    \caption{\small 
Variation in accuracy and token count (\# tok) with adjustments to threshold 
\(T\) on the SQD dataset, with dashed lines indicating oracle accuracy and corresponding token count.}
    \label{fig:oracle}
    \vspace{-0.15in}
\end{figure}

\paragraph{Impact of Threshold Adjustment.}
To examine the impact of varying \( T \), we adjusted the threshold in increments of 10, starting from the 10th percentile, and measured the resulting performance. Additionally, to explore the theoretical maximum performance of our method, we configured an oracle setting where any correct response, regardless of the threshold setting, was counted as correct.

As shown in Figure \ref{fig:oracle}, increasing the threshold \( T \) generally improves performance by removing irrelevant content, thus reducing the number of tokens. However, our experimental results revealed that the performance at the 90th percentile threshold was 29.4, while a lower 80th percentile threshold yielded better performance at 29.9. This indicates that an overly stringent threshold can also remove essential information, suggesting that task-specific threshold fine-tuning could improve results.

Furthermore, in the oracle setting, accuracy significantly improved to 34.1, and the token count was reduced to 77. This shows a marked performance improvement over the best performing threshold (80th percentile), with a similar reduction in tokens. This result implies that dynamically adjusting the threshold based on the query could achieve substantial performance improvements with a comparable number of tokens, suggesting an area for future work. Detailed results are available in Table \ref{tab:detailed_oracle}.

\label{various_threshold}

\begin{figure}[t!]
    \centering
    \includegraphics[width=1\linewidth]{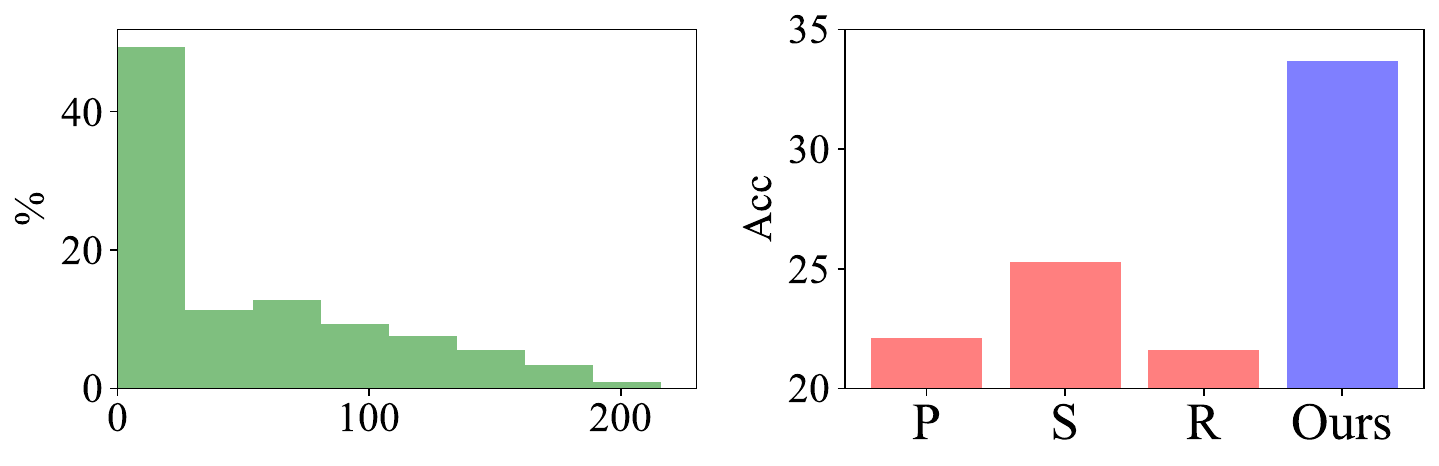}
    \caption{\small (Left) Distribution of token counts in \textit{DSLR}-refined documents on the NQ dataset. (Right) Comparison of \textit{DSLR} with document truncated to an average fixed length (P), document processed using sentence-level re-ranking to include only the most relevant sentences up to the average length (S), and document using random selection of sentences up to the average length (R).}
    \label{fig:distribution}
    \vspace{-0.15in}
\end{figure}
 
\begin{figure}[t!]
    \centering
    \includegraphics[width=1\linewidth]{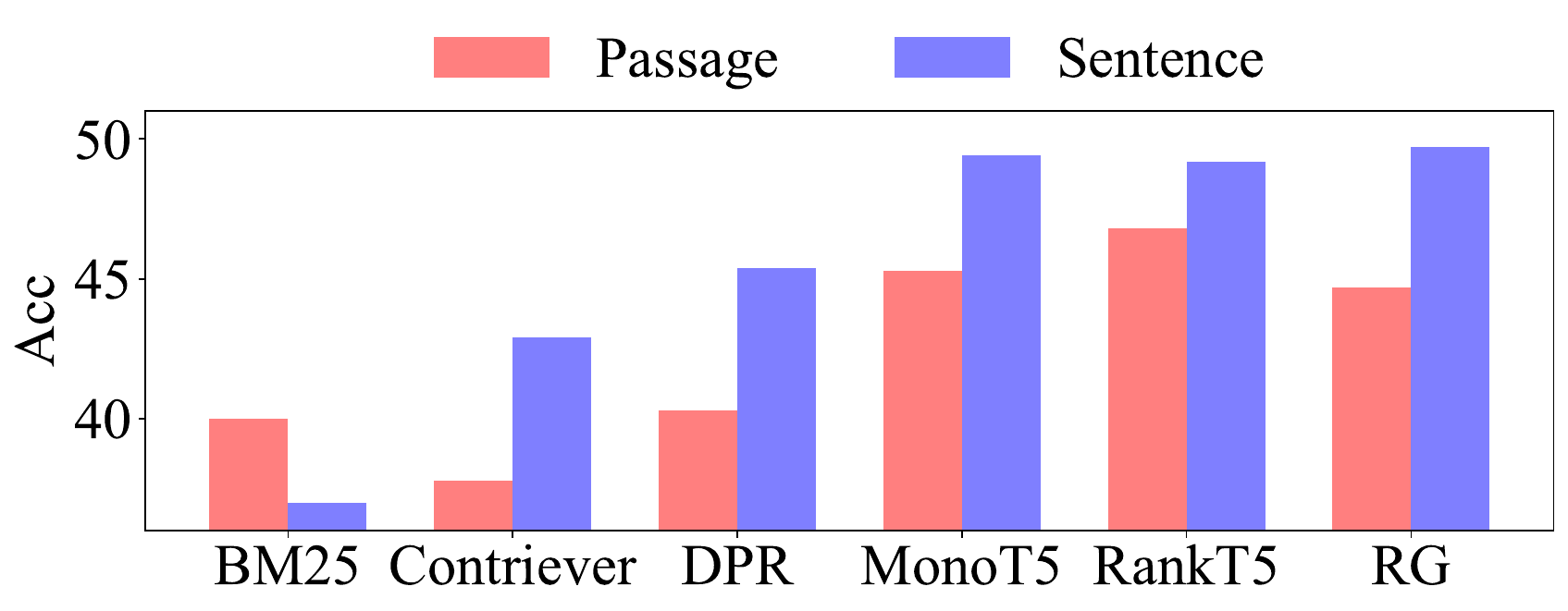}
    \caption{\small Comparative average performance of sentence-level and passage-level re-ranking across the dataset with a context length of 100 words.}
    \label{fig:effective}
    \vspace{-0.15in}
\end{figure}

\paragraph{Token Distribution and Refinement Strategies.}

\begin{figure*}

    \centering
    
    \includegraphics[width=1\linewidth]{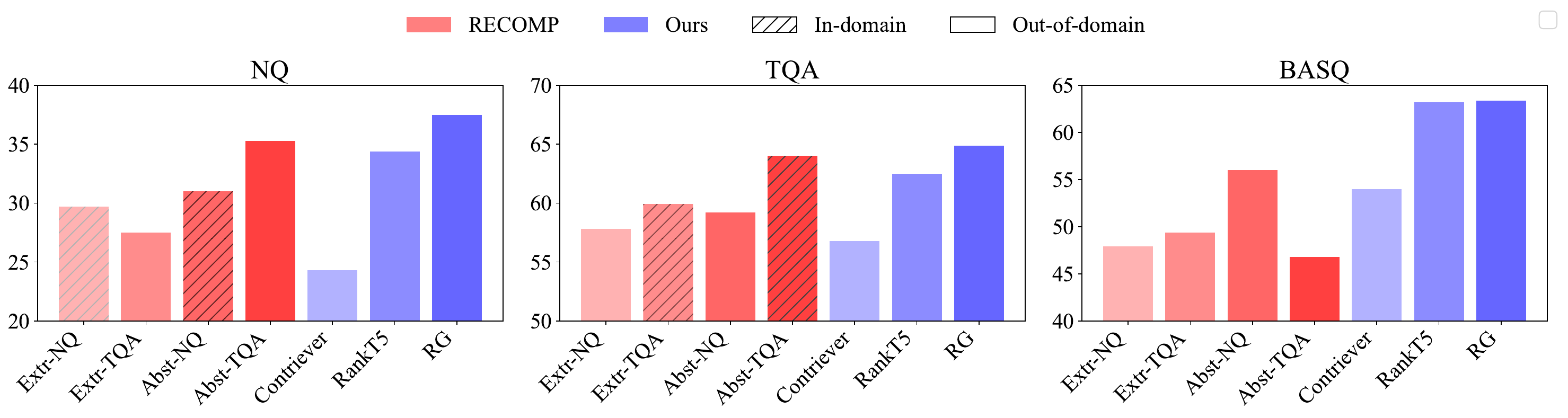}
    
        \caption{\small Performance comparison of \textit{DSLR} and RECOMP \cite{recomp} across multiple open-domain QA datasets, featuring models including Contriever, RankT5, and RG for \textit{DSLR}, and extractive (Extr.) and abstractive (Abst.) models for RECOMP. The in-domain (Hatched) results refer to models specifically trained for the dataset.}
    \label{fig:recomp}
\end{figure*}
The left panel of Figure \ref{fig:distribution} displays the distribution of token counts in documents refined by \textit{DSLR}. Unlike methods that trim passages to a fixed length, \textit{DSLR} reduces token counts based on a relevance score threshold, resulting in a wide distribution of token counts, with many instances nearly devoid of external knowledge. The average token count post-refinement is 46. We analyzed performance by comparing this approach with cases where passages are consistently cut to 46 tokens: one where passages are simply truncated at 46 tokens, another using sentence-level re-ranking to select the most relevant sentences up to 46 tokens, and a third where sentences are randomly cut to 46 tokens.

As demonstrated in the right panel of Figure \ref{fig:distribution}, \textit{DSLR}, which trims content based on relevance, significantly outperforms methods that trim to a fixed length, improving scores from 25.3 to 33.7. This suggests that trimming based on relevance score thresholds, rather than a fixed length, is more effective. This method accommodates the variability in the amount of relevant information per query, indicating that non-essential content should be dynamically removed.

\paragraph{Effectiveness of Sentence-Level Re-ranking.}




\begin{table}[t!]
\small
\captionsetup{font=small, skip=0.1in}
\centering
\begin{tabular}{lccc}
\toprule 
& \textbf{NQ} & \textbf{TQA} \\
\midrule
\textit{DSLR} (Ours) & 33.7 & 64.1 \\
- sentence-level re-ranking & 30.6 & 62.0 \\
- reconstruction (descend) & 33.6 & 63.8 \\
- reconstruction (ascend) & 33.6 & 63.9 \\
- reconstruction (random) & 33.5 & 63.8 \\
\midrule
Baseline & 25.6 & 58.0 \\
\bottomrule

\end{tabular}
\caption{\small Ablation studies on the NQ, TQA datasets, comparing \textit{DSLR} with RG against the variants that exclude sentence-level re-ranking and reconstruction. The variants are ordered by relevance score (descend and ascend) or randomly (random).}

\label{tab:ablation}
\end{table}
To assess the effectiveness of sentence-level re-ranking within our framework, we compared it to conventional passage-level re-ranking using the same context length in RAG, under an initial top-100 retrieval setting. Figure \ref{fig:effective} demonstrates that sentence-level re-ranking markedly outperforms passage-level re-ranking by enhancing performance through increased information density at a finer granularity. Additionally, while dense retrievers and fine-tuned ranking
models demonstrate improvements as re-rankers, BM25 as a re-ranker significantly decreases the performance. This highlights the limitations of lexcial-based retrieval for assessing low-granularity, sentence-level relevance, underscoring the necessity for semantic understanding in sentence ranking tasks.
Moreover, off-the-shelf ranking models, originally designed for passage-level relevance assessment, are also effective at determining relevance at the more granular level of individual sentences. Interestingly, even though it is not specifically trained for ranking tasks, the unsupervised re-ranker using LLMs shows remarkable performance in sentence-level re-ranking.

\paragraph{Ablation Studies on the Sentence-Level Re-ranking and Reconstruction Steps.}
To see how each step in \textit{DSLR} contributes to the overall performance, we conduct the ablation studies, the results shown in Table \ref{tab:ablation}, for the sentence-level re-ranking and reconstruction steps. These studies were uniquely tailored to the variable token counts reduced by \textit{DSLR}, rather than using a fixed length.

First, we examine the impact of removing the sentence-level re-ranking step. In this scenario, after initially retrieving the top-1 passage, the results are decomposed into sentences. Subsequently, these sentences are randomly used as sources for generating answers. The performance drastically drops from 33.7 to 30.6 on the NQ, highlighting the crucial role of sentence-level re-ranking, which helps effectively filter out query-irrelevant information based on relevance scores.

Furthermore, we explore the effectiveness of the reconstruction step. The performance also drops from 64.1 to 63.8 on the TQA. This finding is similar to those from \citet{decontextualization}, which suggests that removing contextual coherence negatively affects the performance. Therefore, in \textit{DSLR}, reconstructing the order of sentences to reflect their original sequence within the retrieved passage is an essential step. Interestingly, the widely used approach of prepending external knowledge in descending order of relevance scores is not effective in our sentence-level refinement framework, showing similar results to a randomly ordered setting.

\newcommand{\hlc}[2][yellow]{{\sethlcolor{#1}\hl{#2}}}
\begin{table*}[ht]
    
    \vspace{-0.1in}
    \centering
    \small
    
    \resizebox{\textwidth}{!}{
    \begin{tabular}{>{\raggedright\arraybackslash}p{3cm}>{\raggedright\arraybackslash}p{6.5cm}>{\raggedright\arraybackslash}p{6.5cm}}
        \toprule
        \textbf{Query} & 
        \textbf{Original Document} & \textbf{DSLR-Refined Document} \\
        \midrule
        the element which is the most abundant in the human body is (NQ) & 
       \multicolumn{1}{p{6.5cm}}{[1] Nitrogen\newline
diatomic gas with the formula N. Dinitrogen forms about 78\% of Earth's atmosphere, making it the most abundant uncombined element. \hlc[red!20]{Nitrogen occurs in all organisms, primarily in amino acids (and thus proteins), in the nucleic acids (DNA and RNA) and in the energy transfer molecule adenosine triphosphate.} \hlc[blue!20]{The human body contains about 3\% nitrogen by mass, the fourth most abundant element in the body after oxygen, carbon, and hydrogen.} \hlc[red!20]{The 
 nitrogen cycle describes movement of the element from the air, into the biosphere and organic compounds, then back into the atmosphere. Many industrially important compounds, such as ammonia, nitric acid,}}
& 
        
\multicolumn{1}{p{6.5cm}}{[1] Nitrogen\newline
diatomic gas with the formula N. Dinitrogen forms about 78\% of Earth's atmosphere, making it the most abundant uncombined element. \hlc[blue!20]{The human body contains about 3\% nitrogen by mass, the fourth most abundant element in the body after oxygen, carbon, and hydrogen.}} \\

        \midrule
        \textbf{Predict} & 
         Nitrogen (\textcolor{red}{X}) & 
        Oxygen (\textcolor{blue}{O})\\
        \bottomrule
    \end{tabular}
}

    \caption{
    \small Case study with the top-1 document, where we represent query-irrelevant sentences in \hlc[red!20]{red} and query-relevant sentences in \hlc[blue!20]{blue}.
    }
    \vspace{-0.125in}
    \label{tab:case_study}
\end{table*}

\paragraph{Comparative Analysis of Document Refining methods: Evaluating RECOMP and \textit{DSLR}.} 
We further compare our \textit{DSLR} to the concurrent supervised refinement method, RECOMP \cite{recomp}, which requires additional training steps for refining the retrieved documents.
To be specific, RECOMP is designed to refine the retrieved passages by either abstractively or extractively summarizing them with additional models.
Note that due to significant differences between supervised and unsupervised schemes, directly comparing \textit{DSLR} with RECOMP on an apples-to-apples basis is difficult.
However, to ensure as fair a comparison as possible, we evaluate both refining methods under the same conditions by adopting a two-sentence extraction context length, following the extractive setting used for RECOMP. 
Additionally, RECOMP's extractive compressor, which requires Contriever to be fine-tuned on specific datasets, shares similarities with our \textit{DSLR} implementation that also uses Contriever, though ours is not additionally fine-tuned.

Figure \ref{fig:recomp} shows the results of the comparison between \textit{DSLR} and RECOMP in both in-domain and out-of-domain settings. While RECOMP shows robust performance on the in-domain datasets where it is particularly trained, its performance drops drastically for the out-of-domain settings, notably for BASQ from 54 to 47.9. This indicates the challenges of dataset-specific tuning for the supervised refinement methods. On the other hand, our \textit{DSLR} with RankT5 and RG shows robust performance even without additional training steps for refinement.

\paragraph{Case Study.}
We conduct a case study of the \textit{DSLR} framework in Table \ref{tab:case_study}. 
Specifically, a conventional fixed-size passage may contain distractors, such as unrelated knowledge and irrelevant conceptual details about Nitrogen (highlighted in red).
Note that, although the retrieved passage-level document includes `Oxygen', which is the correct answer to the given query, the LLM used as the reader fails to generate the accurate answer by being distracted by irrelevant information.
On the other hand, \textit{DSLR} effectively filters out such query-irrelevant sentences. Furthermore, \textit{DSLR} also helps focus on the information closely related to the query (highlighted in blue), thus correctly generating the answer.

\section{Conclusion}



In this work, we present \textit{DSLR}, a novel unsupervised document refinement framework that enhances the performance of RAG systems. The \textit{DSLR} framework aids RAG systems to generate more accurate answers by decomposing passages into sentences, re-ranking them based on each relevance score, and then reconstructing them to preserve the continuity and coherence of the context. Our comprehensive experiments on multiple QA datasets show that \textit{DSLR} consistently outperforms the conventional approaches of using fixed-size passage in RAG, especially in ever-evolving and domain-specific contexts. Our ablation studies highlight the importance of sentence-level re-ranking and contextual reconstruction for improvement on RAG. We believe that \textit{DSLR} suggests a promising research direction for refining document retrieval without additional training, together with potential applications across a wide range of knowledge-intensive NLP tasks by integrating more diverse retrieval or ranking models. 

\normalem

\section*{Limitation}

While our \textit{DSLR} shows significant improvements in RAG performance, it is important to recognize that there is still room for further improvement. 
First, although we aim to preserve the original contextual integrity with the reconstruction step, there is a risk of unintentionally removing important sentences that might contain query-relevant information. We believe that developing more advanced re-ranking models to more accurately capture relevance scores could address this, which we leave as valuable future work.
Second, since \textit{DSLR} aims to refine the set of retrieved documents, there might be a bottleneck stemming from the initial retrieval step; the overall performance can be negatively affected by incorrectly retrieved documents. Therefore, future work may focus on developing a more precise retrieval module.
Since the \textit{DSLR} framework is composed of off-the-shelf modules, we believe that its overall performance will improve concurrently with the development of these modules.

\section*{Ethics Statement}
The experimental results on \textit{DSLR} validate the effectiveness of sentence-level re-ranking and reconstruction in RAG. However, since RAG requires processing a large amount of textual data, we should always be aware of the documents containing sensitive or private information when applying it to real-world scenarios. While it is not within the scope of our study, we believe that developing filtering strategies to mitigate such problems is essential.

\section*{Acknowledgments}

This work was supported by Institute for Information and Communications Technology Promotion (IITP) grant funded by the Korea government (No. 2018-0-00582, Prediction and augmentation of the credibility distribution via linguistic analysis and automated evidence document collection), Basic Science Research Program through the National Research Foundation of Korea (NRF) funded by the Ministry of Education (RS-2023-00275747), and the Artificial Intelligence Industrial Convergence Cluster Development project funded by the Ministry of Science and ICT (MSIT, Korea) \& Gwangju Metropolitan City.



\bibliography{custom}

\begin{thebibliography}{55}
\expandafter\ifx\csname natexlab\endcsname\relax\def\natexlab#1{#1}\fi

\bibitem[{Asai et~al.(2024)Asai, Wu, Wang, Sil, and Hajishirzi}]{selfrag}
Akari Asai, Zeqiu Wu, Yizhong Wang, Avirup Sil, and Hannaneh Hajishirzi. 2024.
\newblock \href {https://openreview.net/forum?id=hSyW5go0v8} {Self-{RAG}: Learning to retrieve, generate, and critique through self-reflection}.
\newblock In \emph{The Twelfth International Conference on Learning Representations}.

\bibitem[{Baek et~al.(2023)Baek, Jeong, Kang, Park, and Hwang}]{kalmv}
Jinheon Baek, Soyeong Jeong, Minki Kang, Jong~C. Park, and Sung~Ju Hwang. 2023.
\newblock \href {https://doi.org/10.18653/V1/2023.EMNLP-MAIN.107} {Knowledge-augmented language model verification}.
\newblock In \emph{Proceedings of the 2023 Conference on Empirical Methods in Natural Language Processing, {EMNLP} 2023, Singapore, December 6-10, 2023}, pages 1720--1736. Association for Computational Linguistics.

\bibitem[{Borgeaud et~al.(2022)Borgeaud, Mensch, Hoffmann, Cai, Rutherford, Millican, van~den Driessche, Lespiau, Damoc, Clark, de~Las~Casas, Guy, Menick, Ring, Hennigan, Huang, Maggiore, Jones, Cassirer, Brock, Paganini, Irving, Vinyals, Osindero, Simonyan, Rae, Elsen, and Sifre}]{DBLP:conf/icml/BorgeaudMHCRM0L22}
Sebastian Borgeaud, Arthur Mensch, Jordan Hoffmann, Trevor Cai, Eliza Rutherford, Katie Millican, George van~den Driessche, Jean{-}Baptiste Lespiau, Bogdan Damoc, Aidan Clark, Diego de~Las~Casas, Aurelia Guy, Jacob Menick, Roman Ring, Tom Hennigan, Saffron Huang, Loren Maggiore, Chris Jones, Albin Cassirer, Andy Brock, Michela Paganini, Geoffrey Irving, Oriol Vinyals, Simon Osindero, Karen Simonyan, Jack~W. Rae, Erich Elsen, and Laurent Sifre. 2022.
\newblock \href {https://proceedings.mlr.press/v162/borgeaud22a.html} {Improving language models by retrieving from trillions of tokens}.
\newblock In \emph{International Conference on Machine Learning, {ICML} 2022, 17-23 July 2022, Baltimore, Maryland, {USA}}, volume 162 of \emph{Proceedings of Machine Learning Research}, pages 2206--2240. {PMLR}.

\bibitem[{Brown et~al.(2020)Brown, Mann, Ryder, Subbiah, Kaplan, Dhariwal, Neelakantan, Shyam, Sastry, Askell, Agarwal, Herbert{-}Voss, Krueger, Henighan, Child, Ramesh, Ziegler, Wu, Winter, Hesse, Chen, Sigler, Litwin, Gray, Chess, Clark, Berner, McCandlish, Radford, Sutskever, and Amodei}]{fewshotlearner}
Tom~B. Brown, Benjamin Mann, Nick Ryder, Melanie Subbiah, Jared Kaplan, Prafulla Dhariwal, Arvind Neelakantan, Pranav Shyam, Girish Sastry, Amanda Askell, Sandhini Agarwal, Ariel Herbert{-}Voss, Gretchen Krueger, Tom Henighan, Rewon Child, Aditya Ramesh, Daniel~M. Ziegler, Jeffrey Wu, Clemens Winter, Christopher Hesse, Mark Chen, Eric Sigler, Mateusz Litwin, Scott Gray, Benjamin Chess, Jack Clark, Christopher Berner, Sam McCandlish, Alec Radford, Ilya Sutskever, and Dario Amodei. 2020.
\newblock \href {https://proceedings.neurips.cc/paper/2020/hash/1457c0d6bfcb4967418bfb8ac142f64a-Abstract.html} {Language models are few-shot learners}.
\newblock In \emph{Advances in Neural Information Processing Systems 33: Annual Conference on Neural Information Processing Systems 2020, NeurIPS 2020, December 6-12, 2020, virtual}.

\bibitem[{Chase(2022)}]{langchain}
Harrison Chase. 2022.
\newblock \href {https://github.com/langchain-ai/langchain} {{LangChain}}.

\bibitem[{Chen et~al.(2023)Chen, Wang, Chen, Yu, Ma, Zhao, Zhang, and Yu}]{denseXretrieval}
Tong Chen, Hongwei Wang, Sihao Chen, Wenhao Yu, Kaixin Ma, Xinran Zhao, Hongming Zhang, and Dong Yu. 2023.
\newblock \href {https://doi.org/10.48550/ARXIV.2312.06648} {Dense {X} retrieval: What retrieval granularity should we use?}
\newblock \emph{arXiv preprint arXiv:2312.06648}, abs/2312.06648.

\bibitem[{Cho et~al.(2023)Cho, Seo, Jeong, and Park}]{das}
Sukmin Cho, Jeongyeon Seo, Soyeong Jeong, and Jong~C. Park. 2023.
\newblock \href {https://doi.org/10.18653/V1/2023.FINDINGS-EMNLP.207} {Improving zero-shot reader by reducing distractions from irrelevant documents in open-domain question answering}.
\newblock In \emph{Findings of the Association for Computational Linguistics: {EMNLP} 2023, Singapore, December 6-10, 2023}, pages 3145--3157. Association for Computational Linguistics.

\bibitem[{Choi et~al.(2021)Choi, Palomaki, Lamm, Kwiatkowski, Das, and Collins}]{decontextualization}
Eunsol Choi, Jennimaria Palomaki, Matthew Lamm, Tom Kwiatkowski, Dipanjan Das, and Michael Collins. 2021.
\newblock \href {https://doi.org/10.1162/TACL\_A\_00377} {Decontextualization: Making sentences stand-alone}.
\newblock \emph{Trans. Assoc. Comput. Linguistics}, 9:447--461.

\bibitem[{Izacard et~al.(2022)Izacard, Caron, Hosseini, Riedel, Bojanowski, Joulin, and Grave}]{contriever}
Gautier Izacard, Mathilde Caron, Lucas Hosseini, Sebastian Riedel, Piotr Bojanowski, Armand Joulin, and Edouard Grave. 2022.
\newblock \href {https://openreview.net/forum?id=jKN1pXi7b0} {Unsupervised dense information retrieval with contrastive learning}.
\newblock \emph{Trans. Mach. Learn. Res.}, 2022.

\bibitem[{Jeong et~al.(2022)Jeong, Baek, Cho, Hwang, and Park}]{dar}
Soyeong Jeong, Jinheon Baek, Sukmin Cho, Sung~Ju Hwang, and Jong~C. Park. 2022.
\newblock \href {https://doi.org/10.18653/V1/2022.ACL-SHORT.48} {Augmenting document representations for dense retrieval with interpolation and perturbation}.
\newblock In \emph{Proceedings of the 60th Annual Meeting of the Association for Computational Linguistics (Volume 2: Short Papers), {ACL} 2022, Dublin, Ireland, May 22-27, 2022}, pages 442--452. Association for Computational Linguistics.

\bibitem[{Jeong et~al.(2024)Jeong, Baek, Cho, Hwang, and Park}]{adaptive-rag}
Soyeong Jeong, Jinheon Baek, Sukmin Cho, Sung~Ju Hwang, and Jong~C. Park. 2024.
\newblock \href {https://doi.org/10.48550/ARXIV.2403.14403} {Adaptive-rag: Learning to adapt retrieval-augmented large language models through question complexity}.
\newblock \emph{arXiv.2403.14403}, abs/2403.14403.

\bibitem[{Jeong et~al.(2023)Jeong, Baek, Hwang, and Park}]{proconvqa}
Soyeong Jeong, Jinheon Baek, Sung~Ju Hwang, and Jong Park. 2023.
\newblock \href {https://doi.org/10.18653/V1/2023.FINDINGS-ACL.374} {Phrase retrieval for open domain conversational question answering with conversational dependency modeling via contrastive learning}.
\newblock In \emph{Findings of the Association for Computational Linguistics: {ACL} 2023, Toronto, Canada, July 9-14, 2023}, pages 6019--6031. Association for Computational Linguistics.

\bibitem[{Jin et~al.(2024)Jin, Zhu, Zhou, and Dou}]{bider}
Jiajie Jin, Yutao Zhu, Yujia Zhou, and Zhicheng Dou. 2024.
\newblock \href {https://doi.org/10.48550/ARXIV.2402.12174} {{BIDER:} bridging knowledge inconsistency for efficient retrieval-augmented llms via key supporting evidence}.
\newblock \emph{arXiv.2402.12174}, abs/2402.12174.

\bibitem[{Joshi et~al.(2017)Joshi, Choi, Weld, and Zettlemoyer}]{TQA}
Mandar Joshi, Eunsol Choi, Daniel~S. Weld, and Luke Zettlemoyer. 2017.
\newblock \href {https://doi.org/10.18653/V1/P17-1147} {Triviaqa: {A} large scale distantly supervised challenge dataset for reading comprehension}.
\newblock In \emph{Proceedings of the 55th Annual Meeting of the Association for Computational Linguistics, {ACL} 2017, Vancouver, Canada, July 30 - August 4, Volume 1: Long Papers}, pages 1601--1611. Association for Computational Linguistics.

\bibitem[{Karpukhin et~al.(2020)Karpukhin, Oguz, Min, Lewis, Wu, Edunov, Chen, and Yih}]{DPR-karpukhin}
Vladimir Karpukhin, Barlas Oguz, Sewon Min, Patrick S.~H. Lewis, Ledell Wu, Sergey Edunov, Danqi Chen, and Wen{-}tau Yih. 2020.
\newblock \href {https://doi.org/10.18653/V1/2020.EMNLP-MAIN.550} {Dense passage retrieval for open-domain question answering}.
\newblock In \emph{Proceedings of the 2020 Conference on Empirical Methods in Natural Language Processing, {EMNLP} 2020, Online, November 16-20, 2020}, pages 6769--6781. Association for Computational Linguistics.

\bibitem[{Kasai et~al.(2023)Kasai, Sakaguchi, Takahashi, Bras, Asai, Yu, Radev, Smith, Choi, and Inui}]{Realtime}
Jungo Kasai, Keisuke Sakaguchi, Yoichi Takahashi, Ronan~Le Bras, Akari Asai, Xinyan Yu, Dragomir Radev, Noah~A. Smith, Yejin Choi, and Kentaro Inui. 2023.
\newblock \href {http://papers.nips.cc/paper\_files/paper/2023/hash/9941624ef7f867a502732b5154d30cb7-Abstract-Datasets\_and\_Benchmarks.html} {Realtime {QA:} what's the answer right now?}
\newblock In \emph{Advances in Neural Information Processing Systems 36: Annual Conference on Neural Information Processing Systems 2023, NeurIPS 2023, New Orleans, LA, USA, December 10 - 16, 2023}.

\bibitem[{Khandelwal et~al.(2020)Khandelwal, Levy, Jurafsky, Zettlemoyer, and Lewis}]{DBLP:conf/iclr/KhandelwalLJZL20}
Urvashi Khandelwal, Omer Levy, Dan Jurafsky, Luke Zettlemoyer, and Mike Lewis. 2020.
\newblock \href {https://openreview.net/forum?id=HklBjCEKvH} {Generalization through memorization: Nearest neighbor language models}.
\newblock In \emph{8th International Conference on Learning Representations, {ICLR} 2020, Addis Ababa, Ethiopia, April 26-30, 2020}. OpenReview.net.

\bibitem[{Krithara et~al.(2023)Krithara, Nentidis, Bougiatiotis, and Paliouras}]{bioasq2}
Anastasia Krithara, Anastasios Nentidis, Konstantinos Bougiatiotis, and Georgios Paliouras. 2023.
\newblock \href {https://doi.org/10.1038/s41597-023-02068-4} {{BioASQ-QA}: A manually curated corpus for biomedical question answering}.
\newblock \emph{Scientific Data}, 10(1):170.
\newblock Published 2023 Mar 27.

\bibitem[{Kwiatkowski et~al.(2019)Kwiatkowski, Palomaki, Redfield, Collins, Parikh, Alberti, Epstein, Polosukhin, Devlin, Lee, Toutanova, Jones, Kelcey, Chang, Dai, Uszkoreit, Le, and Petrov}]{NQ}
Tom Kwiatkowski, Jennimaria Palomaki, Olivia Redfield, Michael Collins, Ankur~P. Parikh, Chris Alberti, Danielle Epstein, Illia Polosukhin, Jacob Devlin, Kenton Lee, Kristina Toutanova, Llion Jones, Matthew Kelcey, Ming{-}Wei Chang, Andrew~M. Dai, Jakob Uszkoreit, Quoc Le, and Slav Petrov. 2019.
\newblock \href {https://doi.org/10.1162/TACL\_A\_00276} {Natural questions: a benchmark for question answering research}.
\newblock \emph{Trans. Assoc. Comput. Linguistics}, 7:452--466.

\bibitem[{Lee et~al.(2021{\natexlab{a}})Lee, Sung, Kang, and Chen}]{phrase_retriever}
Jinhyuk Lee, Mujeen Sung, Jaewoo Kang, and Danqi Chen. 2021{\natexlab{a}}.
\newblock \href {https://doi.org/10.18653/V1/2021.ACL-LONG.518} {Learning dense representations of phrases at scale}.
\newblock In \emph{Proceedings of the 59th Annual Meeting of the Association for Computational Linguistics and the 11th International Joint Conference on Natural Language Processing, {ACL/IJCNLP} 2021, (Volume 1: Long Papers), Virtual Event, August 1-6, 2021}, pages 6634--6647. Association for Computational Linguistics.

\bibitem[{Lee et~al.(2021{\natexlab{b}})Lee, Wettig, and Chen}]{DBLP:conf/emnlp/LeeWC21}
Jinhyuk Lee, Alexander Wettig, and Danqi Chen. 2021{\natexlab{b}}.
\newblock \href {https://doi.org/10.18653/V1/2021.EMNLP-MAIN.297} {Phrase retrieval learns passage retrieval, too}.
\newblock In \emph{Proceedings of the 2021 Conference on Empirical Methods in Natural Language Processing, {EMNLP} 2021, Virtual Event / Punta Cana, Dominican Republic, 7-11 November, 2021}, pages 3661--3672. Association for Computational Linguistics.

\bibitem[{Lewis et~al.(2020)Lewis, Perez, Piktus, Petroni, Karpukhin, Goyal, K{\"{u}}ttler, Lewis, Yih, Rockt{\"{a}}schel, Riedel, and Kiela}]{RAG}
Patrick S.~H. Lewis, Ethan Perez, Aleksandra Piktus, Fabio Petroni, Vladimir Karpukhin, Naman Goyal, Heinrich K{\"{u}}ttler, Mike Lewis, Wen{-}tau Yih, Tim Rockt{\"{a}}schel, Sebastian Riedel, and Douwe Kiela. 2020.
\newblock \href {https://proceedings.neurips.cc/paper/2020/hash/6b493230205f780e1bc26945df7481e5-Abstract.html} {Retrieval-augmented generation for knowledge-intensive {NLP} tasks}.
\newblock In \emph{Advances in Neural Information Processing Systems 33: Annual Conference on Neural Information Processing Systems 2020, NeurIPS 2020, December 6-12, 2020, virtual}.

\bibitem[{Li et~al.(2023)Li, Rawat, Zaheer, Wang, Lukasik, Veit, Yu, and Kumar}]{distractor2}
Daliang Li, Ankit~Singh Rawat, Manzil Zaheer, Xin Wang, Michal Lukasik, Andreas Veit, Felix~X. Yu, and Sanjiv Kumar. 2023.
\newblock \href {https://doi.org/10.18653/V1/2023.FINDINGS-ACL.112} {Large language models with controllable working memory}.
\newblock In \emph{Findings of the Association for Computational Linguistics: {ACL} 2023, Toronto, Canada, July 9-14, 2023}, pages 1774--1793. Association for Computational Linguistics.

\bibitem[{Liang et~al.(2022)Liang, Bommasani, Lee, Tsipras, Soylu, Yasunaga, Zhang, Narayanan, Wu, Kumar, Newman, Yuan, Yan, Zhang, Cosgrove, Manning, R{\'{e}}, Acosta{-}Navas, Hudson, Zelikman, Durmus, Ladhak, Rong, Ren, Yao, Wang, Santhanam, Orr, Zheng, Y{\"{u}}ksekg{\"{o}}n{\"{u}}l, Suzgun, Kim, Guha, Chatterji, Khattab, Henderson, Huang, Chi, Xie, Santurkar, Ganguli, Hashimoto, Icard, Zhang, Chaudhary, Wang, Li, Mai, Zhang, and Koreeda}]{holistic}
Percy Liang, Rishi Bommasani, Tony Lee, Dimitris Tsipras, Dilara Soylu, Michihiro Yasunaga, Yian Zhang, Deepak Narayanan, Yuhuai Wu, Ananya Kumar, Benjamin Newman, Binhang Yuan, Bobby Yan, Ce~Zhang, Christian Cosgrove, Christopher~D. Manning, Christopher R{\'{e}}, Diana Acosta{-}Navas, Drew~A. Hudson, Eric Zelikman, Esin Durmus, Faisal Ladhak, Frieda Rong, Hongyu Ren, Huaxiu Yao, Jue Wang, Keshav Santhanam, Laurel~J. Orr, Lucia Zheng, Mert Y{\"{u}}ksekg{\"{o}}n{\"{u}}l, Mirac Suzgun, Nathan Kim, Neel Guha, Niladri~S. Chatterji, Omar Khattab, Peter Henderson, Qian Huang, Ryan Chi, Sang~Michael Xie, Shibani Santurkar, Surya Ganguli, Tatsunori Hashimoto, Thomas Icard, Tianyi Zhang, Vishrav Chaudhary, William Wang, Xuechen Li, Yifan Mai, Yuhui Zhang, and Yuta Koreeda. 2022.
\newblock \href {https://doi.org/10.48550/ARXIV.2211.09110} {Holistic evaluation of language models}.
\newblock \emph{arXiv preprint arXiv:2211.09110}, abs/2211.09110.

\bibitem[{Liu et~al.(2023)Liu, Lin, Hewitt, Paranjape, Bevilacqua, Petroni, and Liang}]{lostinthemiddle}
Nelson~F. Liu, Kevin Lin, John Hewitt, Ashwin Paranjape, Michele Bevilacqua, Fabio Petroni, and Percy Liang. 2023.
\newblock \href {https://doi.org/10.48550/ARXIV.2307.03172} {Lost in the middle: How language models use long contexts}.
\newblock \emph{arXiv:2307.03172}, abs/2307.03172.

\bibitem[{Mallen et~al.(2023)Mallen, Asai, Zhong, Das, Khashabi, and Hajishirzi}]{DBLP:conf/acl/MallenAZDKH23}
Alex Mallen, Akari Asai, Victor Zhong, Rajarshi Das, Daniel Khashabi, and Hannaneh Hajishirzi. 2023.
\newblock \href {https://doi.org/10.18653/V1/2023.ACL-LONG.546} {When not to trust language models: Investigating effectiveness of parametric and non-parametric memories}.
\newblock In \emph{Proceedings of the 61st Annual Meeting of the Association for Computational Linguistics (Volume 1: Long Papers), {ACL} 2023, Toronto, Canada, July 9-14, 2023}, pages 9802--9822. Association for Computational Linguistics.

\bibitem[{Nogueira and Cho(2019)}]{bert-rerank}
Rodrigo~Frassetto Nogueira and Kyunghyun Cho. 2019.
\newblock \href {http://arxiv.org/abs/1901.04085} {Passage re-ranking with {BERT}}.
\newblock \emph{arXiv, 1901.04085}, abs/1901.04085.

\bibitem[{Nogueira et~al.(2020)Nogueira, Jiang, Pradeep, and Lin}]{monot5}
Rodrigo~Frassetto Nogueira, Zhiying Jiang, Ronak Pradeep, and Jimmy Lin. 2020.
\newblock \href {https://doi.org/10.18653/V1/2020.FINDINGS-EMNLP.63} {Document ranking with a pretrained sequence-to-sequence model}.
\newblock In \emph{Findings of the Association for Computational Linguistics: {EMNLP} 2020, Online Event, 16-20 November 2020}, volume {EMNLP} 2020 of \emph{Findings of {ACL}}, pages 708--718. Association for Computational Linguistics.

\bibitem[{OpenAI(2023{\natexlab{a}})}]{plugin}
OpenAI. 2023{\natexlab{a}}.
\newblock \href {https://openai.com/blog/chatgpt-plugins} {Chatgpt plugins}.

\bibitem[{OpenAI(2023{\natexlab{b}})}]{GPT-4_technical_report}
OpenAI. 2023{\natexlab{b}}.
\newblock \href {https://doi.org/10.48550/arXiv.2303.08774} {{GPT-4} technical report}.
\newblock \emph{arXiv:2303.08774}, abs/2303.08774.

\bibitem[{Petroni et~al.(2021)Petroni, Piktus, Fan, Lewis, Yazdani, Cao, Thorne, Jernite, Karpukhin, Maillard, Plachouras, Rockt{\"{a}}schel, and Riedel}]{KILT}
Fabio Petroni, Aleksandra Piktus, Angela Fan, Patrick S.~H. Lewis, Majid Yazdani, Nicola~De Cao, James Thorne, Yacine Jernite, Vladimir Karpukhin, Jean Maillard, Vassilis Plachouras, Tim Rockt{\"{a}}schel, and Sebastian Riedel. 2021.
\newblock \href {https://doi.org/10.18653/V1/2021.NAACL-MAIN.200} {{KILT:} a benchmark for knowledge intensive language tasks}.
\newblock In \emph{Proceedings of the 2021 Conference of the North American Chapter of the Association for Computational Linguistics: Human Language Technologies, {NAACL-HLT} 2021, Online, June 6-11, 2021}, pages 2523--2544. Association for Computational Linguistics.

\bibitem[{Ponte and Croft(1998)}]{IR}
Jay~M. Ponte and W.~Bruce Croft. 1998.
\newblock \href {https://doi.org/10.1145/290941.291008} {A language modeling approach to information retrieval}.
\newblock In \emph{{SIGIR} '98: Proceedings of the 21st Annual International {ACM} {SIGIR} Conference on Research and Development in Information Retrieval, August 24-28 1998, Melbourne, Australia}, pages 275--281. {ACM}.

\bibitem[{Press et~al.(2023)Press, Zhang, Min, Schmidt, Smith, and Lewis}]{self-ask}
Ofir Press, Muru Zhang, Sewon Min, Ludwig Schmidt, Noah~A. Smith, and Mike Lewis. 2023.
\newblock \href {https://aclanthology.org/2023.findings-emnlp.378} {Measuring and narrowing the compositionality gap in language models}.
\newblock In \emph{Findings of the Association for Computational Linguistics: {EMNLP} 2023, Singapore, December 6-10, 2023}, pages 5687--5711. Association for Computational Linguistics.

\bibitem[{Qin et~al.(2024)Qin, Liang, Ye, Zhu, Yan, Lu, Lin, Cong, Tang, Qian, Zhao, Hong, Tian, Xie, Zhou, Gerstein, dahai li, Liu, and Sun}]{Toolllm}
Yujia Qin, Shihao Liang, Yining Ye, Kunlun Zhu, Lan Yan, Yaxi Lu, Yankai Lin, Xin Cong, Xiangru Tang, Bill Qian, Sihan Zhao, Lauren Hong, Runchu Tian, Ruobing Xie, Jie Zhou, Mark Gerstein, dahai li, Zhiyuan Liu, and Maosong Sun. 2024.
\newblock \href {https://openreview.net/forum?id=dHng2O0Jjr} {Tool{LLM}: Facilitating large language models to master 16000+ real-world {API}s}.
\newblock In \emph{The Twelfth International Conference on Learning Representations}.

\bibitem[{Qin et~al.(2023{\natexlab{a}})Qin, Jagerman, Hui, Zhuang, Wu, Shen, Liu, Liu, Metzler, Wang, and Bendersky}]{llm-rerank}
Zhen Qin, Rolf Jagerman, Kai Hui, Honglei Zhuang, Junru Wu, Jiaming Shen, Tianqi Liu, Jialu Liu, Donald Metzler, Xuanhui Wang, and Michael Bendersky. 2023{\natexlab{a}}.
\newblock \href {https://doi.org/10.48550/ARXIV.2306.17563} {Large language models are effective text rankers with pairwise ranking prompting}.
\newblock \emph{arXiv:2306.17563}, abs/2306.17563.

\bibitem[{Qin et~al.(2023{\natexlab{b}})Qin, Jagerman, Hui, Zhuang, Wu, Shen, Liu, Liu, Metzler, Wang, and Bendersky}]{prp}
Zhen Qin, Rolf Jagerman, Kai Hui, Honglei Zhuang, Junru Wu, Jiaming Shen, Tianqi Liu, Jialu Liu, Donald Metzler, Xuanhui Wang, and Michael Bendersky. 2023{\natexlab{b}}.
\newblock \href {https://doi.org/10.48550/ARXIV.2306.17563} {Large language models are effective text rankers with pairwise ranking prompting}.
\newblock \emph{arXiv:2306.17563}, abs/2306.17563.

\bibitem[{Raffel et~al.(2020)Raffel, Shazeer, Roberts, Lee, Narang, Matena, Zhou, Li, and Liu}]{T5}
Colin Raffel, Noam Shazeer, Adam Roberts, Katherine Lee, Sharan Narang, Michael Matena, Yanqi Zhou, Wei Li, and Peter~J. Liu. 2020.
\newblock \href {http://jmlr.org/papers/v21/20-074.html} {Exploring the limits of transfer learning with a unified text-to-text transformer}.
\newblock \emph{J. Mach. Learn. Res.}, 21:140:1--140:67.

\bibitem[{Rajpurkar et~al.(2016)Rajpurkar, Zhang, Lopyrev, and Liang}]{SQD}
Pranav Rajpurkar, Jian Zhang, Konstantin Lopyrev, and Percy Liang. 2016.
\newblock \href {https://doi.org/10.18653/V1/D16-1264} {Squad: 100, 000+ questions for machine comprehension of text}.
\newblock In \emph{Proceedings of the 2016 Conference on Empirical Methods in Natural Language Processing, {EMNLP} 2016, Austin, Texas, USA, November 1-4, 2016}, pages 2383--2392. The Association for Computational Linguistics.

\bibitem[{Robertson and Zaragoza(2009)}]{bm25}
Stephen~E. Robertson and Hugo Zaragoza. 2009.
\newblock \href {https://doi.org/10.1561/1500000019} {The probabilistic relevance framework: {BM25} and beyond}.
\newblock \emph{Found. Trends Inf. Retr.}, 3(4):333--389.

\bibitem[{Salton and Buckley(1988)}]{tfidf}
Gerard Salton and Chris Buckley. 1988.
\newblock \href {https://doi.org/10.1016/0306-4573(88)90021-0} {Term-weighting approaches in automatic text retrieval}.
\newblock \emph{Inf. Process. Manag.}, 24(5):513--523.

\bibitem[{Seo et~al.(2019)Seo, Lee, Kwiatkowski, Parikh, Farhadi, and Hajishirzi}]{phrase}
Min~Joon Seo, Jinhyuk Lee, Tom Kwiatkowski, Ankur~P. Parikh, Ali Farhadi, and Hannaneh Hajishirzi. 2019.
\newblock \href {https://doi.org/10.18653/V1/P19-1436} {Real-time open-domain question answering with dense-sparse phrase index}.
\newblock In \emph{Proceedings of the 57th Conference of the Association for Computational Linguistics, {ACL} 2019, Florence, Italy, July 28- August 2, 2019, Volume 1: Long Papers}, pages 4430--4441. Association for Computational Linguistics.

\bibitem[{Shi et~al.(2023{\natexlab{a}})Shi, Chen, Misra, Scales, Dohan, Chi, Sch{\"{a}}rli, and Zhou}]{distractor}
Freda Shi, Xinyun Chen, Kanishka Misra, Nathan Scales, David Dohan, Ed~H. Chi, Nathanael Sch{\"{a}}rli, and Denny Zhou. 2023{\natexlab{a}}.
\newblock \href {https://proceedings.mlr.press/v202/shi23a.html} {Large language models can be easily distracted by irrelevant context}.
\newblock In \emph{International Conference on Machine Learning, {ICML} 2023, 23-29 July 2023, Honolulu, Hawaii, {USA}}, volume 202 of \emph{Proceedings of Machine Learning Research}, pages 31210--31227. {PMLR}.

\bibitem[{Shi et~al.(2023{\natexlab{b}})Shi, Min, Yasunaga, Seo, James, Lewis, Zettlemoyer, and Yih}]{replug}
Weijia Shi, Sewon Min, Michihiro Yasunaga, Minjoon Seo, Rich James, Mike Lewis, Luke Zettlemoyer, and Wen{-}tau Yih. 2023{\natexlab{b}}.
\newblock \href {https://doi.org/10.48550/ARXIV.2301.12652} {{REPLUG:} retrieval-augmented black-box language models}.
\newblock \emph{arXiv.2301.12652}, abs/2301.12652.

\bibitem[{Touvron et~al.(2023)Touvron, Martin, Stone, Albert, Almahairi, Babaei, Bashlykov, Batra, Bhargava, Bhosale, Bikel, Blecher, Canton{-}Ferrer, Chen, Cucurull, Esiobu, Fernandes, Fu, Fu, Fuller, Gao, Goswami, Goyal, Hartshorn, Hosseini, Hou, Inan, Kardas, Kerkez, Khabsa, Kloumann, Korenev, Koura, Lachaux, Lavril, Lee, Liskovich, Lu, Mao, Martinet, Mihaylov, Mishra, Molybog, Nie, Poulton, Reizenstein, Rungta, Saladi, Schelten, Silva, Smith, Subramanian, Tan, Tang, Taylor, Williams, Kuan, Xu, Yan, Zarov, Zhang, Fan, Kambadur, Narang, Rodriguez, Stojnic, Edunov, and Scialom}]{Llama2}
Hugo Touvron, Louis Martin, Kevin Stone, Peter Albert, Amjad Almahairi, Yasmine Babaei, Nikolay Bashlykov, Soumya Batra, Prajjwal Bhargava, Shruti Bhosale, Dan Bikel, Lukas Blecher, Cristian Canton{-}Ferrer, Moya Chen, Guillem Cucurull, David Esiobu, Jude Fernandes, Jeremy Fu, Wenyin Fu, Brian Fuller, Cynthia Gao, Vedanuj Goswami, Naman Goyal, Anthony Hartshorn, Saghar Hosseini, Rui Hou, Hakan Inan, Marcin Kardas, Viktor Kerkez, Madian Khabsa, Isabel Kloumann, Artem Korenev, Punit~Singh Koura, Marie{-}Anne Lachaux, Thibaut Lavril, Jenya Lee, Diana Liskovich, Yinghai Lu, Yuning Mao, Xavier Martinet, Todor Mihaylov, Pushkar Mishra, Igor Molybog, Yixin Nie, Andrew Poulton, Jeremy Reizenstein, Rashi Rungta, Kalyan Saladi, Alan Schelten, Ruan Silva, Eric~Michael Smith, Ranjan Subramanian, Xiaoqing~Ellen Tan, Binh Tang, Ross Taylor, Adina Williams, Jian~Xiang Kuan, Puxin Xu, Zheng Yan, Iliyan Zarov, Yuchen Zhang, Angela Fan, Melanie Kambadur, Sharan Narang, Aur{\'{e}}lien Rodriguez, Robert Stojnic, Sergey Edunov,
  and Thomas Scialom. 2023.
\newblock \href {https://doi.org/10.48550/ARXIV.2307.09288} {Llama 2: Open foundation and fine-tuned chat models}.
\newblock \emph{arXiv preprint arXiv:2307.09288}, abs/2307.09288.

\bibitem[{Tsatsaronis et~al.(2015)Tsatsaronis, Balikas, Malakasiotis, Partalas, Zschunke, Alvers, Weissenborn, Krithara, Petridis, Polychronopoulos, Almirantis, Pavlopoulos, Baskiotis, Gallinari, Arti{\`{e}}res, Ngomo, Heino, Gaussier, Barrio{-}Alvers, Schroeder, Androutsopoulos, and Paliouras}]{bioasq}
George Tsatsaronis, Georgios Balikas, Prodromos Malakasiotis, Ioannis Partalas, Matthias Zschunke, Michael~R. Alvers, Dirk Weissenborn, Anastasia Krithara, Sergios Petridis, Dimitris Polychronopoulos, Yannis Almirantis, John Pavlopoulos, Nicolas Baskiotis, Patrick Gallinari, Thierry Arti{\`{e}}res, Axel{-}Cyrille~Ngonga Ngomo, Norman Heino, {\'{E}}ric Gaussier, Liliana Barrio{-}Alvers, Michael Schroeder, Ion Androutsopoulos, and Georgios Paliouras. 2015.
\newblock \href {https://doi.org/10.1186/S12859-015-0564-6} {An overview of the {BIOASQ} large-scale biomedical semantic indexing and question answering competition}.
\newblock \emph{{BMC} Bioinform.}, 16:138:1--138:28.

\bibitem[{Vaswani et~al.(2017)Vaswani, Shazeer, Parmar, Uszkoreit, Jones, Gomez, Kaiser, and Polosukhin}]{attention}
Ashish Vaswani, Noam Shazeer, Niki Parmar, Jakob Uszkoreit, Llion Jones, Aidan~N. Gomez, Lukasz Kaiser, and Illia Polosukhin. 2017.
\newblock \href {https://proceedings.neurips.cc/paper/2017/hash/3f5ee243547dee91fbd053c1c4a845aa-Abstract.html} {Attention is all you need}.
\newblock In \emph{Advances in Neural Information Processing Systems 30: Annual Conference on Neural Information Processing Systems 2017, December 4-9, 2017, Long Beach, CA, {USA}}, pages 5998--6008.

\bibitem[{Wang et~al.(2024)Wang, Ren, Li, Zhao, Liu, and Wen}]{REAR}
Yuhao Wang, Ruiyang Ren, Junyi Li, Wayne~Xin Zhao, Jing Liu, and Ji{-}Rong Wen. 2024.
\newblock \href {https://doi.org/10.48550/ARXIV.2402.17497} {{REAR:} {A} relevance-aware retrieval-augmented framework for open-domain question answering}.
\newblock \emph{arXiv preprint arXiv:2402.17497}, abs/2402.17497.

\bibitem[{Wang et~al.(2019)Wang, Ng, Ma, Nallapati, and Xiang}]{100words}
Zhiguo Wang, Patrick Ng, Xiaofei Ma, Ramesh Nallapati, and Bing Xiang. 2019.
\newblock \href {https://doi.org/10.18653/V1/D19-1599} {Multi-passage {BERT:} {A} globally normalized {BERT} model for open-domain question answering}.
\newblock In \emph{Proceedings of the 2019 Conference on Empirical Methods in Natural Language Processing and the 9th International Joint Conference on Natural Language Processing, {EMNLP-IJCNLP} 2019, Hong Kong, China, November 3-7, 2019}, pages 5877--5881. Association for Computational Linguistics.

\bibitem[{Wang et~al.(2023)Wang, Araki, Jiang, Parvez, and Neubig}]{filco}
Zhiruo Wang, Jun Araki, Zhengbao Jiang, Md.~Rizwan Parvez, and Graham Neubig. 2023.
\newblock \href {https://doi.org/10.48550/ARXIV.2311.08377} {Learning to filter context for retrieval-augmented generation}.
\newblock \emph{arXiv.2311.08377}, abs/2311.08377.

\bibitem[{Welbl et~al.(2017)Welbl, Liu, and Gardner}]{sciq}
Johannes Welbl, Nelson~F. Liu, and Matt Gardner. 2017.
\newblock \href {https://doi.org/10.18653/v1/w17-4413} {Crowdsourcing multiple choice science questions}.
\newblock In \emph{Proceedings of the 3rd Workshop on Noisy User-generated Text, NUT@EMNLP 2017, Copenhagen, Denmark, September 7, 2017}, pages 94--106. Association for Computational Linguistics.

\bibitem[{Wu et~al.(2024)Wu, Shen, and Jiang}]{distractor3}
Zhenyu Wu, Chao Shen, and Meng Jiang. 2024.
\newblock \href {https://doi.org/10.48550/ARXIV.2403.12744} {Instructing large language models to identify and ignore irrelevant conditions}.
\newblock \emph{arXiv.2403.12744}, abs/2403.12744.

\bibitem[{Xu et~al.(2024)Xu, Shi, and Choi}]{recomp}
Fangyuan Xu, Weijia Shi, and Eunsol Choi. 2024.
\newblock \href {https://openreview.net/forum?id=mlJLVigNHp} {{RECOMP}: Improving retrieval-augmented {LM}s with context compression and selective augmentation}.
\newblock In \emph{The Twelfth International Conference on Learning Representations}.

\bibitem[{Yu et~al.(2023{\natexlab{a}})Yu, Iter, Wang, Xu, Ju, Sanyal, Zhu, Zeng, and Jiang}]{gen_read}
Wenhao Yu, Dan Iter, Shuohang Wang, Yichong Xu, Mingxuan Ju, Soumya Sanyal, Chenguang Zhu, Michael Zeng, and Meng Jiang. 2023{\natexlab{a}}.
\newblock \href {https://openreview.net/pdf?id=fB0hRu9GZUS} {Generate rather than retrieve: Large language models are strong context generators}.
\newblock In \emph{The Eleventh International Conference on Learning Representations, {ICLR} 2023, Kigali, Rwanda, May 1-5, 2023}. OpenReview.net.

\bibitem[{Yu et~al.(2023{\natexlab{b}})Yu, Zhang, Liang, Jiang, and Sabharwal}]{ReFeed}
Wenhao Yu, Zhihan Zhang, Zhenwen Liang, Meng Jiang, and Ashish Sabharwal. 2023{\natexlab{b}}.
\newblock \href {https://doi.org/10.48550/ARXIV.2305.14002} {Improving language models via plug-and-play retrieval feedback}.
\newblock \emph{arXiv.2305.14002}, abs/2305.14002.

\bibitem[{Zhuang et~al.(2023)Zhuang, Qin, Jagerman, Hui, Ma, Lu, Ni, Wang, and Bendersky}]{rankt5}
Honglei Zhuang, Zhen Qin, Rolf Jagerman, Kai Hui, Ji~Ma, Jing Lu, Jianmo Ni, Xuanhui Wang, and Michael Bendersky. 2023.
\newblock \href {https://doi.org/10.1145/3539618.3592047} {Rankt5: Fine-tuning {T5} for text ranking with ranking losses}.
\newblock In \emph{Proceedings of the 46th International {ACM} {SIGIR} Conference on Research and Development in Information Retrieval, {SIGIR} 2023, Taipei, Taiwan, July 23-27, 2023}, pages 2308--2313. {ACM}.

\end{thebibliography}
\clearpage

\appendix

\section{Additional Experimental Setups}

\label{sec:Experimental_Setups}

\subsection{Datasets}

\begin{table}[h]
    \centering
    \small
    \begin{tabular}{lc}
        \toprule
        \textbf{Dataset} & \textbf{Number of Queries} \\
        \midrule
        NQ & 8,758 \\
        TQA & 8,837 \\
        SQD & 8,886 \\
        RQA	& 3137 \\
        SQ & 1,000 \\
        BASQ & 1,235 \\
        \bottomrule
    \end{tabular}
    \caption{Detailed number of queries for each dataset used in the experiment.}
    \label{tab:datasets}
\end{table}

Table \ref{tab:datasets} shows the number of queries of the datasets utilized in our experiments. Following \cite{DPR-karpukhin}, we used the development sets of the NQ, TQA, and SQD datasets. The SQ dev-set was also employed. For RQA, we selected answerable queries from documents available on GCS spanning from 2022 to 2023. In BASQ, we selectively employed questions from BioASQ6 challenge (task b) that permitted either factoid or yes/no responses to ensure accuracy.

\subsection{Models}
To construct the retrieval system for our RAG model, we employed BM25 with Pyserini\footnote{\url{https://github.com/castorini/pyserini}}, using pre-indexed corpora provided by the framework. To improve answer generation across datasets, we include document titles to provide context to the LLM, following~\citet{selfrag}. Additionally, recognizing that sentences alone may offer insufficient context, we also included document titles in the reranking process to further ensure contextual richness.

To select models for our re-ranking experiments, we considered a range of realistic scenarios and selected representative models from three key categories: dense retrieval, supervised re-ranking, and unsupervised re-ranking. Specifically, for dense retrieval, we chose DPR and Contriever. In the category of supervised re-ranking, we used the established pointwise ranking models MonoT5 and RankT5. For unsupervised re-ranking, we employed RG, a widely used pointwise re-ranking method. Additionally, acknowledging the significance of latency in practical settings, we favored pointwise methods to efficiently manage the computational overhead associated with processing and decomposing passages into sentences. 

\subsubsection{Model Weights}
All model weights were sourced from Hugging Face, and the models were used without any additional training. Below, we list the specific Hugging Face model names corresponding to the weights employed in our experiments:

\noindent \textbf{DPR}:

- {\small \texttt{facebook/dpr-question\_encoder-multiset-base}}

- {\small \texttt{facebook/dpr-ctx\_encoder-multiset-base}}

\noindent \textbf{Contriever}:

- {\small \texttt{facebook/contriever}}

\noindent \textbf{MonoT5}:

- {\small \texttt{castorini/monot5-base-msmarco}}

\noindent \textbf{RankT5}:

- {\small \texttt{Soyoung97/RankT5-base}}

\noindent \textbf{RECOMP}:

- {\small \texttt{fangyuan/nq\_abstractive\_compressor}}

- {\small \texttt{fangyuan/nq\_extractive\_compressor}}

- {\small \texttt{fangyuan/tqa\_abstractive\_compressor}}

- {\small \texttt{fangyuan/tqa\_extractive\_compressor}}

- {\small \texttt{fangyuan/hotpotqa\_abstractive\_compressor}}

- {\small \texttt{fangyuan/hotpotqa\_extractive\_compressor}}

\noindent\textbf{LLama2-13b-chat}:

- {\small \texttt{meta-llama/Llama-2-13b-chat-hf}}

\begin{figure*}
    \centering
    \includegraphics[width=\linewidth]{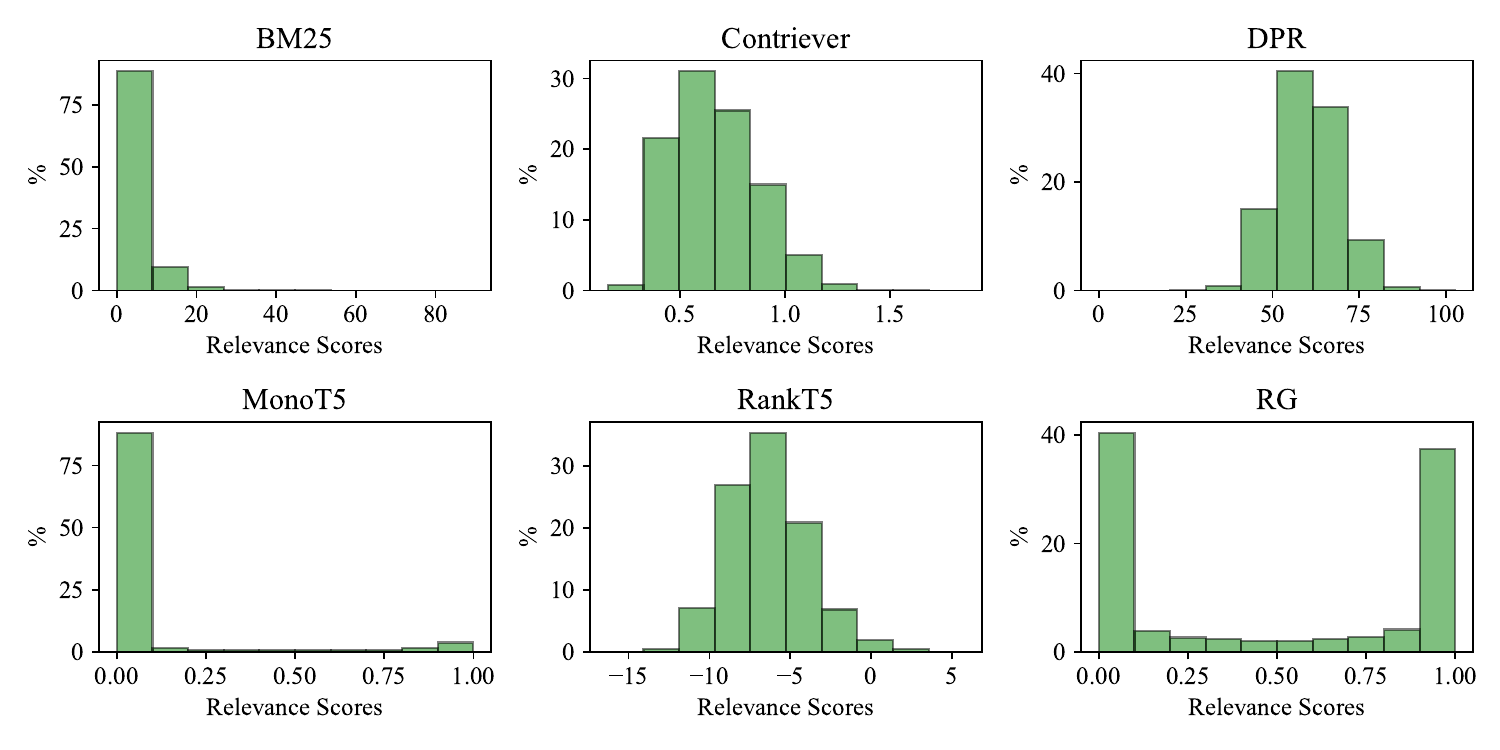}
    \caption{
    Distribution of relevance scores for 1,000 randomly sampled documents from the NQ, TQA, and SQD datasets for each model.
    }
    \label{fig:model_comparison}
\end{figure*}


\begin{table}
    \centering
    
    \small
    \begin{tabular}{lc}
        \toprule
        \textbf{Model} & \multicolumn{1}{c}{\textbf{\( T \)}} \\
        \midrule
        BM25 & 7.6389 \\
        Contriever & 0.9341 \\
        DPR & 71.4338 \\
        MonoT5 & 0.098 \\
        RankT5 & -3.597 \\
        RG & 0.9998 \\
        \bottomrule
    \end{tabular}
    \caption{Threshold \( T \) values used for each model in the main experiments.}
    \label{tab:public_threshold}
\end{table}
\subsubsection{Threshold \(T\) for Each Model}

As shown in the Figure \ref{fig:model_comparison}, the distribution of relevance scores varies significantly across models. Experimentally, we sampled 1,000 entries each from the training sets of the NQ, TQA, and SQD datasets to set the 90th percentile threshold \(T\). Sentences scoring below this threshold were removed. Although it is possible to sample from the training set in each experiment to establish new thresholds, our experiments conducted in Section \ref{various_threshold} across various thresholds consistently yielded better performance than using the top-1 documents directly. Therefore, the thresholds established in this experiment could be used as the standard. The specific values are listed in the accompanying Table \ref{tab:public_threshold}.

\renewcommand{\UrlFont}{\ttfamily\small}
\subsection{Prompt Templates}
For a fair comparison, we fixed the prompt templates. In this section, we introduce these fixed templates. 
\subsubsection{QA Prompt Template}
We use a QA template for open-domain queries from the publicly available llama-index\footnote{\url{https://www.llamaindex.ai/}}. Below is the QA prompt template used in our experiments:\
\begin{tcolorbox}[width=\columnwidth,colback={white},title={QA Prompt Template for LLMs},colbacktitle=white,coltitle=black]

[INST] We have provided context information below.

---------------------

\{context\_str\}

---------------------

Given this information, please answer the question: \{query\_str\} [/INST]
\end{tcolorbox}

\subsubsection{RG Ranking Prompt Template}

We use an RG Ranking Prompt Template following \cite{holistic}. Below is the RG Ranking prompt template used in our experiments:\
\begin{tcolorbox}[width=\columnwidth,colback={white},title={Ranking Prompt Template for LLMs},colbacktitle=white,coltitle=black]

[INST] Passage: 

---------------------

\{title\_str\}

\{document\_str\}

---------------------

Query: \{query\_str\}

Does the passage answer the query? Answer `Yes' or `No' [/INST]
\end{tcolorbox}

\section{Additional Experimental Results}

\subsection{Main Result on Top-5 documents} 

In Table \ref{tab:main5}, we compared the performance of \textit{DSLR}-refined documents for the top-5 settings with original documents in RAG. While \textit{DSLR} remained effective, the margin of performance improvement was less significant than the top-1 setting, suggesting that increasing the volume of documents can modestly enhance performance. However, \textit{DSLR} managed to maintain similar or better performance while significantly reducing token count, thus improving efficiency. In this setting, models like MonoT5, RankT5, and RG, based on pre-trained models, outperformed traditional models such as BM25, Contriever, and DPR, likely due to the superior capability of sentence-level re-ranking.


\begin{table*}[t!]
\vspace{-0.1in}
\small
\centering

\begin{threeparttable}
\renewcommand{\thefootnote}{\fnsymbol{footnote}} 
\resizebox{\textwidth}{!}{%
\begin{tabular}{ll | cc | cc | cc | cc | cc | cc | cc}
\toprule 
\multirow{2}{*}{\textbf{Type}}    & \multirow{2}{*}{\textbf{Re-ranker}}            & \multicolumn{2}{c|}{\textbf{NQ}}            & \multicolumn{2}{c|}{\textbf{TQA}}           & \multicolumn{2}{c|}{\textbf{SQD}}           & \multicolumn{2}{c|}{\textbf{RQA}}           & \multicolumn{2}{c|}{\textbf{SQ}} & \multicolumn{2}{c|}{\textbf{BASQ}} & \multicolumn{2}{c}{\textbf{AVG.}}            \\ 

                              &                    & \# tok& Acc & \# tok& Acc & \# tok& Acc & \# tok& Acc & \# tok& Acc & \# tok& Acc & \# tok& Acc \\ \midrule 
\multicolumn{16}{c}{\textit{Baseline}}   \\  \midrule  \midrule
 - & -   & 855& 34.9& 862& 64.7& 843& 35.9& 2323& 42.5& 825& 40.9& 2131& 62.9& 1307& 47.0\\  \midrule
 \multicolumn{16}{c}{\textit{Ours}}   \\  \midrule \midrule
             Sparse Ret.                &                   BM25                &    204& 29.9& 371& 62.1& 172& 31.6& 1590& 42.6& 231& 39.6& 868& 58.7& 573& 44.1\\ \midrule
\multirow{2}{*}{Dense Ret.}       & {Contriever} 
                                                 & 303& 32.6& 251& 63.8& 243& 34.5& 1163& 44.1& 301& 41.7& 1433& 62.4& 616& 46.6\\ 
                              &  DPR          & 262& \textbf{37.9}& 305& 65.6& 225& 32.8& 1095& 42.5& 334& \textbf{42.3}& 1390& 60.5& 602& 46.8\\ \midrule
\multirow{2}{*}{Supervised Re-r.}   & MonoT5      & 325& 36.1& 353& 65.0& 273& 36.7& 1194& \textbf{44.9}& 200& 40& 1640& 62.3& 664& 47.5\\
                              & RankT5           & 368& 35.8& 285& 64.9& 369& \textbf{36.9}& 976& 44.1& 202& 40.6& 1458& 63.0& 610& 47.6\\ \midrule
Unsupervised Re-r.  & RG  & 198& 37.4& 320& \textbf{66.8}& 205& 35.3& 1099& 43.9& 453& 40.8& 1253& \textbf{63.4}& 588& \textbf{47.9}\\ \bottomrule
\end{tabular}%
}


\end{threeparttable}

\caption{\small Performance comparison between the \textit{Baseline} (original top-5 document) and \textit{Ours} (\textit{DSLR}-refined top-5 document) on various open-domain QA datasets. The table shows the average token count (\# tok) and accuracy (Acc) for both sparse and dense retrievers, as well as for supervised and unsupervised re-rankers. Best results are in \textbf{bold}.}
\label{tab:main5}
\end{table*}

\subsection{Detailed Results for the Comparative Analysis of Document Refining Methods: Evaluating RECOMP and \textit{DSLR}}

\begin{table*}
    \centering
    \small
    \begin{tabular}{cc |cccccc |c}
    \toprule
         \textbf{Method} & \textbf{Model} &  \textbf{NQ} &  \textbf{TQA} &  \textbf{SQD} &  \textbf{RQA} &  \textbf{SQ} &  \textbf{BASQ} & \textbf{AVG.}\\ \midrule
         \multirow{6}*{\textit{DSLR}} &  BM25&  22.2&  53&  27.5&  36.5&  33.7&  54& 37.8\\
         &  DPR&  35&  62&  28.8&  32.9&  38.1&  55.6& 42.1\\
         &  Contriever&  24.3&  56.8&  28.5&  34.9&  37.4&  54& 39.3\\
         &  MonoT5&  34.1&  62.2&  38.9&  40.8 &  37.8 &  60.4 & 45.7\\
         &  RankT5&  34.4&  62.5&  38.9&  42.7&  38.4&  63.2& 46.7\\
         &  RG&  37.5&  64.9&  35.5&  41.4&  41.4&  63.4& \textbf{47.4}\\ \midrule
         \multirow{5}*{RECOMP}&  Extr.-NQ&  29.7&  57.8&  26&  28.2&  38.6&  47.9& 38.0\\
         &  Extr.-TQA&  27.5&  59.9&  27.6&  32.1&  36.7&  49.4& 38.9\\
         &  Extr.-HQA&  27.2&  57.7&  30.3&  33.3&  35.7&  50.9& 39.2\\ 
 \citep{recomp} & Abst.-NQ& 31& 59.2& 34.1& 38.5& 36.1& 56&42.5\\
 & Abst.-TQA& 35.3& 64& 29.2& 37.3& 45.1& 46.8&43.0\\
 & Abst.-HQA& 30.9& 58.3& 33.7& 37.6& 39.9& 41.8&40.4\\ \bottomrule
    \end{tabular}
    \caption{
Performance comparison of \textit{DSLR} and RECOMP methods across multiple open-domain QA datasets. The table presents the accuracy of each method, including BM25, DPR, Contriever, MonoT5, RankT5, and RG models for \textit{DSLR}, as well as extractive (Extr.) and abstractive (Abst.) models for RECOMP. The best performance is in \textbf{bold}.}
    \label{tab:recomp}
\end{table*}

Table \ref{tab:recomp} provides a detailed comparison between the RECOMP and \textit{DSLR} frameworks. RECOMP focuses on minimizing token usage in RAG without sacrificing performance, utilizing a fine-tuned Contriever for extractive compression and a T5-large for abstractive compression. By contrast, \textit{DSLR} enhances RAG performance by eliminating redundant content. Although their different objectives pose a challenge for direct comparison, both aim to extract essential information effectively. To ensure a fair comparison, we aligned the context length to two sentences and refined the top-5 documents, mirroring RECOMP’s methodology. Our experiments utilized the LLama2-13b-chat model as the reader to maintain consistency. This analysis underscores the importance of zero-shot refinement approaches in advancing document refinement for RAG.

\begin{table}
    \small
    \centering
    \begin{tabular}{cccc}
        \toprule
        & \textbf{\# tok} & \textbf{gpt-3.5-turbo} & \textbf{claude-3-haiku} \\
        \midrule
        Baseline & 170 & 33.5 & 33.8 \\
        Ours & 47 & 42.9 & 40.8 \\
        \bottomrule
    \end{tabular}
    \caption{\small
Performance comparison of the baseline (original top-1 document) and Ours (\textit{DSLR}-refined top-1 document using RG) on the NQ dataset within proprietary models. The comparison includes average token count (\# tok) and accuracy.}
\vspace{-0.125in}
    \label{tab:gpt}
\end{table}

\subsection{\textit{DSLR} with Proprietary Models} We evaluated the performance of \textit{DSLR} in proprietary LLMs with larger parameter sizes and undisclosed data and training processes, specifically testing on GPT-3.5-turbo\footnote{gpt-3.5-turbo-0125} and Claude-3-haiku\footnote{claude-3-haiku-20240307} using the same settings for the top-1 document. As shown in Table \ref{tab:gpt}, consistent with previous findings, \textit{DSLR} significantly enhanced performance by simply eliminating irrelevant content at the sentence level from the original document. Additionally, since these models calculate API costs on a per-token basis, the substantial reduction in token count\footnote{Due to the unavailability of the tokenizers for gpt-3.5-turbo and claude-3-haiku, token counts were necessarily performed using the LlamaTokenizer.} is expected to significantly decrease API costs.

\subsection{Sentence-Level Re-ranking Results}
\label{sec:main1_20}
In \textit{DSLR}, the sentence-level re-ranking step is crucial for enhancing performance. We evaluated this approach against conventional passage-level re-ranking within the RAG framework, maintaining identical context lengths (\(L\)). Initial retrievals were configured for top-\{20, 100\}, followed by analyses at \(L=\{100, 500\}\). These settings were chosen because 100 and 500 words represent typical lengths for segments in top-1 and top-5 passage-level re-rankings, respectively. Notably, when counting words, only the content is considered, excluding titles.
\subsubsection{Comparative Performance of Sentence-Level vs. Passage-Level Re-Ranking}

The results presented in Table \ref{tab:main1_20} demonstrate that sentence-level re-ranking consistently outperforms passage-level re-ranking across all settings, except when using BM25.

\label{top_20_results}


\begin{table*}[t!]

\vspace{-0.1in}
\small
\centering

\begin{threeparttable}
\renewcommand{\thefootnote}{\fnsymbol{footnote}} 
\resizebox{\textwidth}{!}{%
\begin{tabular}{lll | cc | cc | cc | cc | cc | cc | cc}
\toprule 
\multirow{2}{*}{\textbf{Type}}    & \multirow{2}{*}{\textbf{Re-ranker}}             & \multirow{2}{*}{\textbf{Granularity}} & \multicolumn{2}{c|}{\textbf{NQ}}            & \multicolumn{2}{c|}{\textbf{TQA}}           & \multicolumn{2}{c|}{\textbf{SQD}}           & \multicolumn{2}{c|}{\textbf{RQA}\tnote{*}}           & \multicolumn{2}{c|}{\textbf{SQ}} & \multicolumn{2}{c|}{\textbf{BASQ}} & \multicolumn{2}{c}{\textbf{AVG.}}            \\ 

                              &                                    &    & \textit{L}=100 & \textit{L}=500 & \textit{L}=100 & \textit{L}=500 & \textit{L}=100 & \textit{L}=500 & \textit{L}=100 & \textit{L}=500 & \textit{L}=100 & \textit{L}=500 & \textit{L}=100 & \textit{L}=500 & \textit{L}=100 & \textit{L}=500 \\ \midrule
\multicolumn{17}{c}{\textit{w/o Re-ranking}}   \\  \midrule
- & - & -  & 25.5  & {34.9}  & 58    & {64.6}  & 28.5  & {35.9}  & 37    & {40.1}  & 33.9  & {40.9}  & 56.8  & {59.5} & 40.0 & 46.0 \\ \midrule
\multicolumn{17}{c}{\textit{w/ Re-ranking}}   \\  \midrule

\multicolumn{17}{c}{\textit{Top-20}}   \\  \midrule
             Sparse Ret.                 &                   BM25                & Sentence                    & 26.3  & 37.5  & 56.7  & 65.6  & 31.5  & 37.1  & 39.3  & 43.1  & 35.6  & 43.4  & 58.5  & 64.2 & 41.3 & 48.5  \\ \midrule
\multirow{5}{*}{Dense Ret.}       & \multirow{2}{*}{ Contriever} & Passage                     & 26.5  & {37.7}  & 57.5  & {66.1}  & 26.2  & {37.2}  & 33.6  & {38}    & 35.4  & {43.7}  & 53.9  & {57.9} & 38.9 & 46.8  \\
                              &                                    & Sentence                    & 28.5  & {37.2}  & 60.9  & {67}    & 32.9  & {38.4}  & 38.3  & {42.7}  & 39    & {44}    & 60.7  & {63.4} & 43.4 & 48.8  \\ \cmidrule{2-17} 
                              & \multirow{2}{*}{ DPR}  & Passage                     & 36.5  & {42.1}  & 62.3  & {67.3}  & 25.3  & {35.4}  & 31.5  & {36}    & 38.9  & {44.6}  & 52.5  & {56.5} & 41.2 & 47.0  \\
                              &                                    & Sentence                    & 38.5  & {42.5}  & 64.3  & {68.2}  & 33.1  & {36}    & 35.8  & {40.4}  & 41.4  & {45.6}  & 59    & {62.7} & 45.4 & 49.2  \\ \midrule
\multirow{5}{*}{Supervised Rer.}   & \multirow{2}{*}{ MonoT5}     & Passage                     & 33.6  & {40.7}  & 62.1  & {67.6}  & 37.1  & {40}    & 37.8  & {41.5}  & 37.4  & {44.7}  & 58.7  & {62.3} & 44.5 & 49.5 \\
                              &                                    & Sentence                    & 37.4  & {42.3}  & 64.9  & {68.2}  & 39.5  & {39.5}  & 42.1  & {44.6}  & 41.4  & {44}    & 64.2  & {65.1} & 48.3 & 50.6  \\ \cmidrule{2-17} 
                              & \multirow{2}{*}{ RankT5}     & Passage                     & 35.4  & {41.4}  & 63.3  & {67.7}  & 39.2  & {40}    & 37.9  & {40.8}  & 37.8  & {44.5}  & 59.2  & {63.0} & 45.5 & 49.6    \\
                              &                                    & Sentence                    & 36.9  & {42.1}  & 64    & {67.9}  & 39.9  & {39.5}  & 42.8  & {43.9}  & 40.2  & {45.7}  & 65.6  & {66.5} & 48.2 & 50.9  \\ \midrule
\multirow{2}{*}{Unsupervised Rer.} & \multirow{2}{*}{ RG}         & Passage                     & 35.9  & {41.9}  & 65.7  & {68.8}  & 34    & {39.3}  & 27    & {30.1}  & 40.2  & {45}    & 60.9  & {63.5} & 44.0 & 48.1  \\
                              &                                    & Sentence                    & 39.2  & {42.9}  & 67.1  & {68.8}  & 37.2  & {39.8}  & 41.5  & {44.6}  & 42.1  & {47.3}  & 64.7  & {66.9} & \textbf{48.6} & \textbf{51.7}                                                                                                                                                                                                                                                                                                                        
\\ \midrule
\multicolumn{17}{c}{\textit{Top-100}}   \\  \midrule
             Sparse Ret.                &                   BM25                & Sentence  &    22.1  & {33.5}  & 50.1  & {62.0}    & 26.5  & {33.6}  & 39.3  & {43.1}  & 32.2  & {41.2}  & 51.5  & {60.4} & 37.0 & 45.6  \\ \midrule
\multirow{5}{*}{Dense Ret.}       & \multirow{2}{*}{Contriever} & Passage                     & 26.0    & {37.2}  & 56.4  & {66.0}    & 23.7  & {35.3}  & 33.6  & {38.0}    & 36.3  & {44.6}  & 51.0    & {56} & 37.8 & 46.2 \\
                              &                                    & Sentence                    & 28.0    & {37.6}  & 59.4  & {67.5}  & 32.5  & {39.0}    & 38.3  & {42.7}  & 40.5  & {45.4}  & 58.7  & {63.5} & 42.9 & 49.3   \\ \cmidrule{2-17} 
                              & \multirow{2}{*}{DPR}  & Passage                     & 39.2  & {46.5}  & 61.8  & {68.8}  & 22.8  & {33.4}  & 31.5  & {36}    & 38.5  & {44.5}  & 48.0    & {53.6} & 40.3 & 47.1  \\
                              &                                    & Sentence                    & 41.9  & {46.7}  & 65.3  & {69.8}  & 31.8  & {38.0}    & 35.8  & {40.4}  & 40.7  & {48.1}  & 57.0    & {62.1} & 45.4 & 50.9  \\ \midrule
\multirow{5}{*}{Supervised Re-r.}   & \multirow{2}{*}{MonoT5}     & Passage                     & 35.4  & {43.9}  & 62.8  & {69.1}  & 38.3  & {42.3}  & 37.8  & {41.5}  & 39.3  & {46.8}  & 58.4  & {63.2} & 45.3 & 51.1 \\
                              &                                    & Sentence                    & 40.5  & {46.3}  & 65.8  & {70.3}  & 41.9  & {41.8}  & 42.1  & {44.6}  & 42.2  & {48.6}  & 64.0    & {68.0} & 49.4 & 53.3   \\ \cmidrule{2-17} 
                              & \multirow{2}{*}{RankT5}     & Passage                     & 38.0    & {44.7}  & 64.5  & {70.0}    & 41.5  & {43.5}  & 37.9  & {40.8}  & 39.0    & {46.5}  & 59.8  & {64.2} & 46.8 & 51.6  \\
                              &                                    & Sentence                    & 39.7  & {46.0}    & 65.5  & {69.9}  & 42.4  & {41.8}  & 42.8  & {43.9}  & 39.0    & {48.5}  & 65.6  & {68.5} & 49.2 & 53.1  \\ \midrule
\multirow{2}{*}{Unsupervised Re-r.} & \multirow{2}{*}{RG}         & Passage                     & 37.6  & {44.9}  & 66.3  & {71.0}    & 33.5  & {40.8}  & 27.0    & {30.1}  & 40.2  & {46.9}  & 60.8  & {63.4} & 44.2 & 49.5  \\
                              &                                    & Sentence                    & 41.7  & {47.4}  & 68.5  & {71.7}  & 37.5  & {41.5}  & 41.5  & {44.6}  & 43.8  & {49.4}  & 65.4  & {68.8} & \textbf{49.7} & \textbf{53.9}  \\ \bottomrule
\end{tabular}%
}
\begin{tablenotes}
\item[* RQA uses a specific GCS document from the dataset instead of the top-100, allowing for a variable number of top-\(N\) retrieved documents.] 
\end{tablenotes}
\vspace{-0.125in}
\end{threeparttable}
\caption{Comparative performance of sentence-level and passage-level re-ranking methods across multiple open-domain QA datasets. Results are presented for two context lengths (\(L\)=100 and \(L\)=500), using sparse and dense retrievers, and both supervised and unsupervised re-rankers, for the top-{20, 100} retrieved documents. The best performances are in \textbf{bold}.}
\label{tab:main1_20}
\end{table*}

\begin{table*}[ht]
    \centering
    \small

    \begin{tabular}{@{}lc|c|c|c|c|c@{}}
        \toprule
        Re-ranker & Granularity & \(L\)=100 & \(L\)=200 & \(L\)=300 & \(L\)=400 & \(L\)=500 \\
        \midrule
        Contriever & Passage & 26 & 29.9 & 33 & 35.4 & 37.2 \\ 
                   & Sentence & 28 & 32.8 & 35.3 & 36.4 & 37.6 \\
        \midrule
        DPR        & Passage & 39.2 & 42.5 & 44 & 45.8 & 46.5 \\
                   & Sentence & 41.9 & 44.5 & 45.8 & 46.4 & 46.7 \\
        \midrule
        MonoT5     & Passage & 35.4 & 39 & 41.6 & 43 & 43.9 \\
                   & Sentence & 40.5 & 43.9 & 45.6 & 46.1 & 46.3 \\
        \midrule
        RankT5     & Passage & 38 & 41 & 42.6 & 43.9 & 44.7 \\
                   & Sentence & 39.7 & 43.3 & 44.9 & 46.2 & 46 \\
        \midrule
        RG         & Passage & 37.6 & 41 & 42.7 & 44 & 44.9 \\
                   & Sentence & 41.7 & 44.7 & 46.2 & 47.3 & 47.4 \\
        \midrule
        \midrule
        AVG.       & Passage & 35.2 & 38.7 & 40.8 & 42.4 & 43.4 \\
                   & Sentence & 38.4 & 41.8 & 43.6 & 44.5 & 44.8 \\
        \bottomrule
    
    \end{tabular}
    
    \caption{Performance comparison across different context lengths (\(L\) = 100, 200, 300, 400, and 500) on the NQ dataset, evaluated using top-100 retrieved documents.}
    \label{tab:l_variants}
\end{table*}

\begin{table*}[ht]
    \centering
    \small
    \resizebox{\linewidth}{!}{
    \begin{tabular}{llcc|cc|cc|cc|cc}
        \toprule
        \textbf{Re-ranker} & \textbf{Granularity} & \multicolumn{2}{c}{\textbf{Top-5}} & \multicolumn{2}{c}{\textbf{Top-10}} & \multicolumn{2}{c}{\textbf{Top-20}} & \multicolumn{2}{c}{\textbf{Top-50}} & \multicolumn{2}{c}{\textbf{Top-100}} \\
        \cmidrule(lr){3-4} \cmidrule(lr){5-6} \cmidrule(lr){7-8} \cmidrule(lr){9-10} \cmidrule(lr){11-12}
          &  & \(L\)=100 & \(L\)=500 & \(L\)=100 & \(L\)=500 & \(L\)=100 & \(L\)=500 & \(L\)=100 & \(L\)=500 & \(L\)=100 & \(L\)=500 \\
        \midrule
         Contriever & Passage & 26.7 & 35.3 & 26.7 & 37.0 & 26.5 & 37.7 & 26.4 & 37.1 & 26.0 & 37.2 \\
         & Sentence & 28.0 & 34.8 & 27.9 & 36.3 & 28.5 & 37.2 & 28.1 & 37.7 & 28.0 & 37.6 \\ \midrule
         DPR & Passage & 32.8 & 35.9 & 34.5 & 39.5 & 36.5 & 42.1 & 38.5 & 45.0 & 39.2 & 46.5 \\
         & Sentence & 33.0 & 34.9 & 36.3 & 39.0 & 38.5 & 42.5 & 40.9 & 45.6 & 41.9 & 46.7 \\ \midrule
         MonoT5 & Passage & 31.3 & 35.0 & 32.6 & 38.5 & 33.6 & 40.7 & 34.6 & 43.3 & 35.4 & 43.9 \\
         & Sentence & 32.9 & 34.8 & 35.3 & 38.9 & 37.4 & 42.3 & 39.7 & 44.7 & 40.5 & 46.3 \\ \midrule
         RankT5 & Passage & 32.1 & 35.5 & 33.8 & 38.4 & 35.4 & 41.4 & 37.0 & 43.8 & 38.0 & 44.7 \\
         & Sentence & 32.5 & 34.9 & 34.9 & 38.7 & 36.9 & 42.1 & 38.8 & 44.8 & 39.7 & 46.0 \\ \midrule
         RG & Passage & 33.0 & 35.3 & 34.8 & 39.6 & 33.6 & 41.9 & 34.7 & 44.2 & 35.2 & 44.9 \\
         & Sentence & 33.9 & 34.8 & 36.7 & 39.6 & 36.1 & 42.9 & 37.7 & 45.8 & 38.4 & 47.4 \\ \midrule
         \midrule
         AVG. & Passage & 31.2 & 35.4 & 32.5 & 38.6 & 33.6 & 40.8 & 34.7 & 42.7 & 35.2 & 43.4 \\
         & Sentence & 32.1 & 34.8 & 34.2 & 38.5 & 36.1 & 41.4 & 37.7 & 43.7 & 38.4 & 44.8 \\
        \bottomrule
    \end{tabular}}
    \caption{Performance comparison of various re-rankers at different granularity levels and context lengths (\(L\)=100 and \(L\)=500), evaluated on NQ dataset across top-\{5, 10, 20, 50, 100\} retrieved documents.}
    \label{tab:n_variants}
\end{table*}

\subsubsection{Effectiveness of Sentence-Level Re-Ranking in Varying Conditions}
Table \ref{tab:l_variants} shows the sentence-level and passage-level re-ranking over various context lengths \(L\). Table \ref{tab:n_variants} shows performance in top-\{5, 10, 20, 50, 100\} settings adjusted for \(L=100\) and \(L=500\). Our experiments on the NQ dataset indicate that sentence-level re-ranking is effective across diverse conditions, omitting the less effective BM25 re-ranking.

\label{sec:robustness}

\subsubsection{Effectiveness of Sentence-Level Re-Ranking on the Gold Answer Hit Rate}
\label{sec:hit_rate}

We present detailed results for the Gold Answer Hit Rate in Table \ref{tab:hit_rate}. The rate is binary, assigned 1 if the re-ranked context contains the gold answer, and 0 otherwise, averaged over all dataset entries for each \(L\).



\begin{table*}[t!]

\small
\centering

\begin{threeparttable}
\renewcommand{\thefootnote}{\fnsymbol{footnote}} 
\resizebox{\textwidth}{!}{%
\begin{tabular}{lll  |cc  |cc  |cc  |cc  |cc  |cc  |cc}
\toprule 
\multirow{2}{*}{\textbf{Type}}    & \multirow{2}{*}{\textbf{Re-ranker}}             & \multirow{2}{*}{\textbf{Granularity}} & \multicolumn{2}{c|}{\textbf{NQ}}            & \multicolumn{2}{c|}{\textbf{TQA}}           & \multicolumn{2}{c|}{\textbf{SQD}}           & \multicolumn{2}{c|}{\textbf{RQA}\tnote{*}}           & \multicolumn{2}{c|}{\textbf{SQ}} & \multicolumn{2}{c|}{\textbf{BASQ}} & \multicolumn{2}{c}{\textbf{AVG.}}            \\ 

                              &                                    &    & \textit{L}=100 & \textit{L}=500 & \textit{L}=100 & \textit{L}=500 & \textit{L}=100 & \textit{L}=500 & \textit{L}=100 & \textit{L}=500 & \textit{L}=100 & \textit{L}=500 & \textit{L}=100 & \textit{L}=500 & \textit{L}=100 & \textit{L}=500 \\ \midrule

\multicolumn{17}{c}{\textit{Top-20}}   \\ \midrule
\multirow{5}{*}{Dense Ret.}       & \multirow{2}{*}{ Contriever} & Passage                     & 24.1& 49.5& 44.8& 70.5& 29.3& 53.7& 30.3& 42.8& 30.1& 58.9& 21& 32.2& 29.9& 51.3\\
                              &                                    & Sentence                    & 25.9& 47.4& 50.1& 71.7& 39.8& 56.3& 34.8& 48.6& 37.6& 61.4& 19.6& 32.1& 34.6& 52.9\\ \cmidrule{2-17} 
                              & \multirow{2}{*}{ DPR}  & Passage                     & 41.9& 57.7& 59.1& 73.4& 29.5& 52.3& 29.2& 42.1& 37.3& 59.3& 20.6& 30.5& 36.2& 52.6\\
                              &                                    & Sentence                    & 46.6& 59.0& 64.3& 74.6& 42.4& 58.2& 32.8& 49.5& 44.6& 61.8& 24.2& 34.2& 42.5& 56.2\\ \midrule
\multirow{5}{*}{Supervised Rer.}   & \multirow{2}{*}{ MonoT5}     & Passage                     & 37.7& 57.1& 59& 74.1& 45.7& 60.1& 36.1& 47.2& 38.7& 60.4& 25.9& 36.0& 40.5& 55.8\\
                              &                                    & Sentence                    & 46.2& 58.3& 65.6& 74.4& 54.2& 61.5& 43.9& 52.6& 46.5& 63.1& 29.7& 39.2& 47.7& 58.2\\ \cmidrule{2-17} 
                              & \multirow{2}{*}{ RankT5}     & Passage                     & 40.7& 57.9& 61& 74.1& 48.3& 60.7& 34.8& 45.6& 39.9& 61.9& 26.6& 35.9& 41.9& 56\\
                              &                                    & Sentence                    & 44.8& 57.9& 64.5& 73.8& 54.2& 61.7& 43.4& 52.3& 44.9& 63& 30.5& 39.4& 47& 58\\ \midrule
\multirow{2}{*}{Unsupervised Rer.} & \multirow{2}{*}{ RG}         & Passage                     & 38.1& 57.7& 59.9& 74.7& 40.3& 58.9& 21.1& 30.9& 40.5& 61.6& 25.8& 35.1& 37.6& 53.2\\
                              &                                    & Sentence                    & 47.1& 59.3& 66.1& 75.5& 50.8& 61.6& 38.6& 51.4& 48.8& 65.8& 29.4& 38.9& 46.8& 58.7\\ \midrule
\multicolumn{17}{c}{\textit{Top-100}}  \\ \midrule
\multirow{5}{*}{Dense Ret.}       & \multirow{2}{*}{Contriever} & Passage                     & 23& 48.9& 42& 70.2& 25.8& 51.8& 30.3& 42.8& 29.5& 59.7& 19.8& 30.6& 28.4& 50.7\\
                              &                                    & Sentence                    & 24.7& 46.4& 48.2& 70.8& 39.1& 57.7& 34.8& 48.6& 38.6& 63.9& 18.7& 31.6& 34& 53.2\\ \cmidrule{2-17} 
                              & \multirow{2}{*}{DPR}  & Passage                     & 46.5& 64.9& 59.3& 75.3& 26.9& 49.2& 29.2& 42.1& 35.2& 61.6& 20.1& 29.5& 36.2& 53.8\\
                              &                                    & Sentence                    & 52.4& 66.6& 65.9& 77.2& 41.4& 59.5& 32.8& 49.5& 45.1& 67.8& 24.3& 33.4& 43.6& 59\\ \midrule
\multirow{5}{*}{Supervised Re-r.}   & \multirow{2}{*}{MonoT5}     & Passage                     & 40.2& 63& 60.1& 76.8& 48& 65.8& 36.1& 47.2& 41.5& 65.2& 25.7& 36.8& 41.9& 63.6\\
                              &                                    & Sentence                    & 51.1& 66.2& 67.8& 77.9& 60.1& 69.6& 43.9& 52.6& 49.4& 68.2& 29.4& 39.9& 50.3& 62.4\\ \cmidrule{2-17} 
                              & \multirow{2}{*}{RankT5}     & Passage                     & 44.1& 64.2& 62.9& 77.1& 51.6& 67.3& 34.8& 45.6& 41.8& 65.5& 27.4& 36.8& 43.8& 59.4\\
                              &                                    & Sentence                    & 49.8& 65& 66.8& 76.9& 60& 69.4& 43.4& 52.3& 49& 67& 31& 40.5& 50& 61.8\\ \midrule
\multirow{2}{*}{Unsupervised Re-r.} & \multirow{2}{*}{RG}         & Passage                     & 40& 63.1& 60.6& 77.7& 40.1& 61.4& 21.1& 30.9& 40.9& 66.7& 26.5& 35.1& 38.2& 55.8\\
                              &                                    & Sentence                    & 51.2& 66.6& 67.7& 79& 52.8& 67.6& 38.6& 51.4& 54& 71.4& 29.6& 39.8& 49& 62.6\\ \bottomrule
\end{tabular}%
}
\begin{tablenotes}
\item[* RQA uses a specific GCS document from the dataset instead of the top-20, allowing for a variable number of top-\(N\) retrieved documents.] 
\end{tablenotes}
\vspace{-0.1in}
\end{threeparttable}
\caption{Golden Answer Hit rate of sentence-level and passage-level re-ranking methods across multiple open-domain QA datasets. Results are presented for two context lengths (\(L\)=100 and \(L\)=500), using dense retrievers, and both supervised and unsupervised re-rankers, for the top-\{20, 100\} retrieved documents.}
\label{tab:hit_rate}
\end{table*}

\subsubsection{Ablation Studies on Various Models}

\begin{table*}
    \centering
    \small
    \begin{tabular}{cc |ccc|c}
    \toprule
    & \textbf{Model} & {\textbf{NQ}} & {\textbf{TQA}} & {\textbf{SQD}}  & \textbf{AVG.} \\
    \midrule
    \multirow{5}{*}{Sentence-Level Re-ranking} & Contriever & 37.6 & 67.5 & 39.1  & 48.1  \\
                                 & DPR        & 46.7 & 69.9 & 38.1 & 51.6  \\
                                 & MonoT5     & 46.4 & 70.4 & 41.9 & 52.9 \\
                                 
                                 & RankT5     & 46.1 & 70.0 & 41.9 &52.7 \\
                                 & RG         & 47.4 & 71.7 & 41.5 & 53.5 \\
    \midrule
    w/o SR                       &            & 24.1 & 51.0 & 14.4 & 29.8 \\ 
    \midrule
    \multirow{5}{*}{w/o RC (descend)} & Contriever & 36.8 & 66.7 & 38.1 & 47.2\\
                                      & DPR        & 46.9 & 69.3 & 37.6 & 51.3\\
                                      & MonoT5     & 45.9 & 68.9 & 41.6 & 52.1 \\
                                      & RankT5     & 46.0 & 69.3 & 41.3 & 52.5\\
                                      & RG         & 46.3 & 71.0 & 39.6 & 52.3\\
    \midrule
    \multirow{5}{*}{w/o RC (random)} & Contriever & 37.4 & 66.8 & 37.7 & 47.3\\
                                     & DPR        & 46.5 & 69.0 & 37.2 & 50.9\\
                                     & MonoT5     & 46.0 & 70.0 & 40.6 & 52.2\\
                                     & RankT5     & 45.6 & 69.1 & 40.3 & 51.7\\
                                     & RG         & 46.3 & 71.2 & 39.7 &52.4\\
    \bottomrule
    \end{tabular}
    \caption{\small Ablation studies on the NQ, TQA, and SQD datasets comparing the Sentence-Level Re-ranking performance with its variants. This includes the baseline RG model and variants without sentence-level re-ranking (w/o SR) and without reconstruction (w/o RC), evaluated in conditions with scores ordered by relevance (descend) and shuffled randomly (random).}
    \label{tab:ablations}
\end{table*}
\label{sec:ablations}

Table \ref{tab:ablations} explores the significance of each step under various models in the initial top-100 retrieval and \(L\)=500 setting. The absence of the sentence-level re-ranking (SR) highlights its necessity in filtering irrelevant information, while excluding the reconstruction (RC) step demonstrates its crucial role in enhancing answer generation accuracy.

\begin{table*}
    \centering
    \small
    \begin{tabular}{@{}c|ccccc@{}}
        \toprule
        \textbf{\(N\)} & \textbf{1} & \textbf{3} & \textbf{5} & \textbf{7} & \textbf{10} \\
        \midrule
        \multicolumn{6}{c}{\textit{Baseline}} \\
        \midrule
        \midrule
        Acc & 25.6 & 31.7 & 34.9 & 37.0 & 39.6 \\
        \# tok & 169 & 512 & 855 & 1198 & 1713 \\
        E2E  & 3.368 & 4.436 & 5.239 & 6.030 & 7.382 \\
        \midrule
        \multicolumn{6}{c}{\textit{Ours}} \\
        \midrule
        \midrule
        Acc  & 31.1 & 34.0 & 36.1 & 37.6 & 39.7 \\
        \# tok & 74 & 207 & 325 & 431 & 577 \\
        E2E  & 3.792 & 4.081 & 4.232 & 4.559 & 5.422 \\
        \bottomrule
    \end{tabular}
    \caption{Performance comparison at various \(N\)-values for Baseline vs. Ours, using Accuracy (Acc), Token count (\# tok), and End-to-End latency (E2E) on the NQ dataset.}
    \label{tab:detail_N}
\end{table*}

\begin{table*}
    \centering
    
    \small
    
    \begin{tabular}{@{}l|cccccccccc@{}}
        \toprule
        \textbf{{(\%)}} & \textbf{{10}} & \textbf{{20}} & \textbf{{30}} & \textbf{{40}} & \textbf{{50}} & \textbf{{60}} & \textbf{{70}} & \textbf{{80}} & \textbf{{90}} & \textbf{{Oracle}} \\
        \midrule
        \(T\) & 2.7969e-05 & 0.00043 & 0.0076 & 0.0826 & 0.65841 & 0.9196 & 0.9857 & 0.9981 & 0.9998 & - \\
        Acc & 28.6 & 28.7 & 29.0 & 29.2 & 29.4 & 29.7 & 29.8 & 29.9 & 29.5 & 34.1 \\
        \# tok & 164 & 159 & 150 & 141 & 123 & 109 & 94 & 75 & 51 & 77 \\
        \bottomrule
    \end{tabular}
    
    \caption{Variation in accuracy and token count (\# tok) with adjustments to relevance score percentiles, including the set threshold values \( T \) and oracle settings on the NQ dataset.}
    \label{tab:detailed_oracle}
\end{table*}


\end{document}